\documentclass[sigconf]{acmart}
\usepackage{enumitem}
\usepackage{booktabs} % 必须：用于专业三线表
\usepackage{graphicx} % 必须：用于 \resizebox
\usepackage{multirow}
\usepackage{amsmath}
\usepackage[table]{xcolor}

\usepackage{amssymb}
\usepackage{amsthm}

\usepackage{makecell}
\usepackage{pifont}

\usepackage{caption}
\usepackage{framed}
\usepackage{tcolorbox}
\usepackage{xurl} % 允许 URL 在任意字符处换行

\AtBeginDocument{%
  }

\copyrightyear{2026}
\acmYear{2026}
\setcopyright{cc}
\setcctype{by}
\acmConference[MM '26] {Proceedings of the 34th ACM International Conference on Multimedia}{November 10--14, 2026}{Rio de Janeiro, Brazil.}
\acmBooktitle{Proceedings of the 34th ACM International Conference on Multimedia (MM '26), November 10--14, 2026, Rio de Janeiro, Brazil}
\acmISBN{979-8-4007-2213-4/2026/11}
\acmDOI{10.1145/3767308.3835113}

\begin{document}

\title{Local Margin Restoration for Test-Time Adaptation of Vision-Language Models}


\author{Yan Huang}
\authornote{Equal contribution.}
\affiliation{%
  \institution{Guangzhou University}
  \city{Guangzhou}
  \country{China}
}
% \affiliation{%
%   \institution{South China University of Technology}
%   \city{Guangzhou}
%   \country{China}
% }
\email{dennishuangyan@gmail.com}

\author{Guowei Wang}
\authornotemark[1]
\affiliation{%
  \institution{The Second Affiliated Hospital of Guangzhou University of Chinese Medicine}
  \city{Guangzhou}
  \country{China}}
\email{gw.wang@gzucm.edu.cn}

\author{Xu Wang}
\affiliation{%
  \institution{ Jinan University}
  \city{Zhuhai}
  \country{China}
}
\email{xuwang@jnu.edu.cn}

\author{Kangjun Liu}
\affiliation{%
 \institution{Pengcheng Laboratory}
 \city{Shenzhen}
 \country{China}}
\email{liukj@pcl.ac.cn}

\author{Xin Lin}
\correspondingauthor
\authornote{Corresponding author.}
\affiliation{%
  \institution{Guangzhou University}
  \city{Guangzhou}
  \country{China}}
\email{linxin94@gzhu.edu.cn}

% \renewcommand{\shortauthors}{Huang et al.}

\begin{abstract}
Vision-language models (VLMs) such as CLIP exhibit remarkable zero-shot capabilities, yet their performance frequently degrades sharply under unexpected test-time distribution shifts. 
While Test-Time Adaptation (TTA) offers a promising solution, continuously adapting VLMs over an unlabeled test stream presents fundamental challenges.
Conventional top-1-centric updates often reinforce errors by corrupting the local semantic geometry among related classes, while iterative adaptation exacerbates progressive bias accumulation, ultimately driving the model toward mode collapse.
To overcome these coupled vulnerabilities, we propose Local Margin Restoration (LMR), a lightweight, one-step TTA framework. At the sample level, our Protected Margin Restoration (PMR) objective recovers local semantic geometry by shielding plausible near-top candidates from external hard negatives. Concurrently, to combat stream-level degradation, we introduce a dual-stage stabilization mechanism, featuring an Adaptive Margin (AM) controller and Bias Correction (BC), to dynamically disrupt progressive bias accumulation and prevent mode collapse. Extensive experiments on CIFAR-C, ImageNet-C, and ImageNet variants demonstrate that LMR consistently outperforms state-of-the-art TTA baselines, proving exceptionally robust and efficient even in challenging low-batch test-time regimes. Our code is available at \url{https://github.com/DennisHuangYan/LMR}.
\end{abstract}

\begin{CCSXML}
<ccs2012>
   <concept>
       <concept_id>10010147.10010257.10010258.10010262.10010277</concept_id>
       <concept_desc>Computing methodologies~Transfer learning</concept_desc>
       <concept_significance>500</concept_significance>
       </concept>
 </ccs2012>
\end{CCSXML}

\ccsdesc[500]{Computing methodologies~Transfer learning}

\keywords{Vision-Language Models, Test-Time Adaptation}

\maketitle

\section{Introduction}

% Vision-Language Models (VLMs), most notably CLIP \cite{radford21clip}, have demonstrated remarkable success in zero-shot image classification and cross-modal understanding by learning well-aligned representations from massive image-text pairs. Despite their impressive capabilities, deploying these pre-trained models in real-world, dynamic environments poses a significant challenge: their performance could degrade significantly when encountering unexpected test-time distribution shifts~\cite{a2025clipartt,maharana2025batclip}, where corrupted or out-of-distribution inputs distort the visual evidence used for image-text matching. To mitigate this vulnerability without requiring access to the original training data or expensive offline re-training, Test-Time Adaptation (TTA) has emerged as a practical and highly promising paradigm \cite{wang2021tent, niu22Efficient, yuan2023robust}. TTA aims to continuously adapt the model on the fly using only streaming, unlabeled test data.

Vision-Language Models (VLMs), most notably CLIP \cite{radford21clip}, have demonstrated remarkable success in zero-shot image classification and cross-modal understanding by learning well-aligned representations from massive image-text pairs. Despite these impressive capabilities, deploying pre-trained VLMs in dynamic, real-world environments presents a significant challenge: their performance could degrade sharply under unexpected test-time distribution shifts \cite{a2025clipartt,maharana2025batclip}, where corrupted or out-of-distribution inputs distort the visual evidence necessary for accurate image-text matching. To mitigate this vulnerability without requiring access to the original training data or expensive offline re-training, Test-Time Adaptation (TTA) has emerged as a practical and promising paradigm \cite{wang2021tent, niu22Efficient, yuan2023robust,su2022revisiting,Su_Xu_Jia_2024,huang2026enhancing}. TTA aims to continuously adapt the model on the fly using only a stream of unlabeled test data.

Recent studies have extended TTA to VLMs using generic entropy objectives, pseudo-label self-training, and specialized vision-language updates. Despite encouraging progress, the current TTA methods~\cite{a2025clipartt, maharana2025batclip, sun2026bilateral, baomint,osowiechi2024watt} for VLMs remain fundamentally challenging. Many existing approaches seek to improve model confidence by sharpening the predictive distribution either implicitly or explicitly. While effective when the current prediction is reliable, such updates become highly detrimental under severe domain shifts where the pseudo labels are often incorrect~\cite{maharana2025batclip}. Unlike standard classifiers, VLM predictions rely on relative similarities to a fixed set of text prototypes, blindly sharpening these predictions destroys the fine-grained ranking structure among semantically related classes. Consequently, performing adaptation can reinforce mistakes rather than correct them, leading to an accumulation of errors over an extended unlabeled stream.

To understand the fundamental bottlenecks of TTA in VLMs, we conduct an in-depth empirical investigation into the failure modes of zero-shot predictions under severe domain shift, revealing degradation across two coupled dimensions. First, at the sample-level geometry, we observe that many shifted errors remain locally recoverable: the ground-truth class frequently resides within a plausible near-top candidate set, separated from the incorrect top-1 prediction by a remarkably narrow similarity gap. Standard entropy minimization aggressively penalizes this neighborhood, compressing the local margin against external hard negatives and prematurely discarding recoverable knowledge. Second, at the stream level, this vulnerability is compounded over time. Pre-trained VLMs exhibit inherent output biases that favor specific classes. Successive confident updates on these biased pseudo-labels create a destructive feedback loop, amplifying progressive bias accumulation and ultimately driving the model toward total class collapse.

Motivated by these findings, we argue that effective TTA in VLMs must adhere to two core design principles: it must preserve plausible near-top candidates rather than forcing overconfident commitment to the top-1 prediction, and it must establish dynamically calibrated margins to disrupt long-term bias accumulation. To realize these principles, we propose a novel one-step TTA framework: \textbf{Local Margin Restoration (LMR)}. Rather than globally sharpening a single pseudo-label, LMR strategically targets the recoverable local-neighborhood geometry. Specifically, we introduce a {Protected Margin Restoration (PMR)} objective. For each sample, PMR dynamically constructs a protected candidate set encompassing plausible near-top classes. Instead of treating all nearby alternatives as negatives, PMR explicitly restores the semantic separation only against hard external competitors outside this set, effectively shielding the potential ground truth from premature suppression.

Furthermore, to satisfy our second design principle and combat progressive bias accumulation across the continuous data stream, we complement PMR with a dual-stage stabilization strategy. This includes an update-side {Adaptive Margin (AM)} controller that dynamically weakens the target optimization margin for historically over-represented classes before gradients enter the model, and an output-side {Bias Correction (BC)} module that calibrates residual concentration from the emitted predictions. This combined mechanism guarantees that the model resists long-term prediction collapse during test-time adaptation.

% To overcome these fundamental limitations, we propose a novel one-step TTA framework: \textbf{Local Margin Restoration (LMR)}. Rather than globally sharpening a single pseudo-label, our approach strategically targets the recoverable local-neighborhood regime. Specifically, we introduce a \textbf{Protected Margin Restoration (PMR)} objective. For each sample, PMR dynamically constructs a protected candidate set encompassing plausible near-top classes based on similarity scores. Instead of immediately treating all nearby alternatives as negatives, PMR explicitly restores the semantic separation \textit{only} against hard external competitors outside the protected set. By doing so, it effectively shields the potentially correct class from premature suppression and maintains a healthy local semantic geometry.

% Furthermore, to combat the gradual concentration of predictions over an extended online stream, we complement PMR with a dual-stage stream-level stabilization strategy. This includes an update-side \textbf{Adaptive Margin (AM)} controller that dynamically weakens the target optimization margin for over-reinforced pseudo-classes before gradients enter the model, and an output-side \textbf{Bias Correction (BC)} module that calibrates residual concentration from the emitted predictions. This combined mechanism guarantees that the model resists long-horizon prediction collapse during continuous adaptation.

In summary, the main contributions are summarized as follows:
% \begin{itemize}
\begin{itemize}[leftmargin=*, noitemsep, topsep=1pt, parsep=0pt, partopsep=0pt]
    % \item We expose the critical flaws of relying strictly on top-1 confidence for VLM adaptation under severe corruptions, revealing that plausible near-top classes are often discarded prematurely and model predictions tend to suffer from stream-level collapse.

    \item We provide a rigorous failure analysis of TTA in VLMs under severe corruptions, exposing the geometric vulnerability of near-top candidate classes and the progressive stream-level bias accumulation that leads to mode collapse.

    % \item We propose the \textbf{LMR} framework, featuring the \textbf{PMR} objective, which protects near-top candidate classes while strictly pushing away hard external distractors to recover local semantic margins.

    \item We propose \textbf{LMR}, featuring the PMR objective, which safely protects near-top candidate classes while strictly pushing away hard external distractors to recover local semantic margins.
    
    % \item We introduce a lightweight dual-stage stabilization mechanism, comprising an \textbf{AM} controller and a \textbf{BC} module, to regulate class reinforcement and calibrate output bias, ensuring robust continuous adaptation.

    \item We introduce a lightweight dual-stage stabilization mechanism, comprising an AM controller and a BC module, to dynamically regulate class reinforcement and calibrate output bias, ensuring robust long-term adaptation.
    
    % \item Extensive experiments on standard corruption benchmarks (CIFAR-10-C, CIFAR-100-C, and ImageNet-C) demonstrate that our proposed method significantly outperforms state-of-the-art TTA approaches for VLMs, especially in challenging low-batch regimes.

    \item Extensive experiments on corruption benchmarks (CIFAR-10-C, CIFAR-100-C, and ImageNet-C) and ImageNet-Variants benchmarks demonstrate that our proposed method significantly outperforms state-of-the-art TTA approaches for VLMs, especially in challenging low-batch regimes.
\end{itemize}

\section{Related Works}

\noindent\textbf{Test-time adaptation.}
Test-time adaptation adapts a pretrained model during inference using only unlabeled target data. While conventional TTA methods rely on entropy minimization~\cite{wang2021tent}, recent approaches have substantially improved performance through confidence based filtering~\cite{niu22Efficient}, sharpness aware updates~\cite{niu2023towards}, memory replay~\cite{yuan2023robust}, and augmentation-based objectives~\cite{zhang2022memo}. Despite their success in mitigating distribution shifts, these unimodal-centric approaches are sub-optimal for vision-language models, as their underlying design paradigms cannot accommodate the crucial bimodal representation structure, text prototypes, and image-text alignment mechanisms.

\noindent\textbf{Test-time adaptation in vision-language models.}
To overcome the limitations of unimodal approaches, recent works have tailored TTA to CLIP-style vision-language models along several directions. 
Pioneer approaches adapt the learnable prompt at the textual branch, as in TPT~\cite{shu2022tpt} and its variants~\cite{wang2024scp,yoon2024ctpt,zehao2025dynaprompt,sheng2025r,sharifdeen2025tpt}. However, their reliance on time-consuming data augmentation introduces prohibitive latency during inference. Bypassing this bottleneck, another training-free line refines predictions using test-time feature statistics~\cite{karmanov2024efficient}, caches~\cite{zhang2024dual}, or label propagation~\cite{li2025efficient}. While highly efficient, they lack the capacity to handle severe distribution shifts.
Recent works seek an optimal trade-off, focusing on efficient gradient updates at the visual branch. They primarily advance adaptation through weight averaging~\cite{osowiechi2024watt}, multimodal-specific objectives~\cite{maharana2025batclip,baomint,sun2026bilateral,lafoncliptta}, and class-prior correction~\cite{zhou2025bayesian,zanella2025realistic}.

\section{Methodology}

\begin{figure*}[t]
    \centering
    \includegraphics[width=\textwidth]{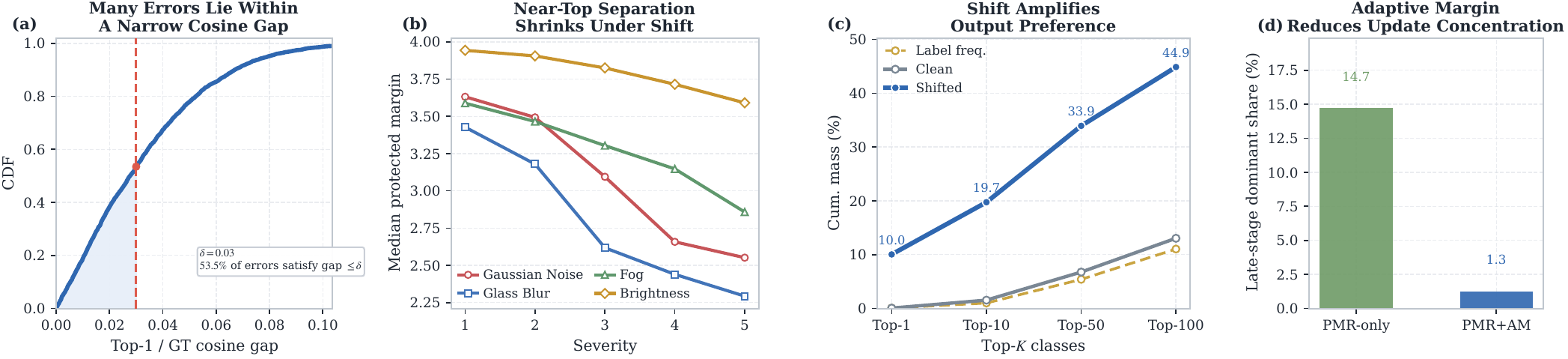}
    \vspace{-0.5cm}
    \caption{\footnotesize{
Empirical failure patterns under test-time shift that motivate our design. Frost at severity 5 is used as a representative shift in (a), (c), and (d), while (b) summarizes representative corruption trends.
(a) Many shifted errors remain locally recoverable under a narrow top-1/GT gap criterion.
(b) Near-top separation progressively shrinks as corruption severity increases.
(c) Shifted predictions become markedly more concentrated than both clean predictions and stream composition.
(d) Local restoration alone leads to strong update concentration, which adaptive margin substantially suppresses.}
}
\Description{A four-panel figure labeled (a) through (d) illustrating failure patterns under test-time shift. Panel (a) is a cumulative distribution function line plot showing that many shifted errors remain locally recoverable under a narrow top-1/GT gap criterion.. Panel (b) is a line plot showing median protected margin versus corruption severity from 1 to 5 across four corruptions, where margins drop sharply as severity increases. Panel (c) is a line plot of cumulative probability mass across top-K classes, showing shifted predictions are heavily concentrated up to 44.8 percent in top-100 compared to clean predictions and label frequency. Panel (d) is a bar chart comparing late-stage dominant class share between PMR-only and PMR+AM, demonstrating that adaptive margin significantly reduces update concentration.}
\label{fig:obs_overview}
\end{figure*}

\subsection{Problem Formulation}
\label{sec:preliminaries}

\paragraph{Zero-Shot CLIP Classification.}
Given a pre-trained CLIP model comprising an image encoder $f_{\theta}$ and a text encoder $f_{\phi}$, we construct the zero-shot classifier for a label space $\mathcal{Y}=\{1,\dots,C\}$. Specifically, we wrap each class name $c \in \mathcal{Y}$ in a task-specific prompt template $T(c)$ (e.g., \textit{``a photo of a [CLASS].''}) and extract its $\ell_2$-normalized text prototype:

% Consider a pre-trained Vision-Language Model (VLM), such as CLIP, comprising an image encoder $f_{\theta}$ and a text encoder $f_{\phi}$.
% Given a classification task with label space $\mathcal{Y}=\{1,\dots,C\}$, we construct a discrete text prototype for each class $c \in \mathcal{Y}$. Specifically, we wrap each class name in a task-specific prompt template $T(c)$ (e.g., \textit{``a photo of a [CLASS].''}) and extract its $\ell_2$-normalized text embedding:
\begin{equation}
\textbf{w}_c = \frac{f_{\phi}(T(c))}{\|f_{\phi}(T(c))\|_2}  \in \mathbb{R}^d,
\end{equation}
where $d$ is the embedding dimension. Once constructed, the text encoder and these prototypes are kept frozen.
% These class embeddings are stacked to form the global semantic bank $\mathbf{W}=[\textbf{w}_1,\dots,\textbf{w}_C]^\top \in \mathbb{R}^{C\times d}$,  which serves as the weights for the zero-shot classifier.
% The text encoder $f_{\phi}$ and the resulting semantic bank $\mathbf{W}$ are kept frozen during the online testing phase.
 For an input image $\mathbf{x}_i$, the visual encoder extracts the normalized feature $\mathbf{z}_i = f_{\theta}(\mathbf{x}_i) \in \mathbb{R}^d$. We first compute the cosine similarity $s_{i,c}$ between the visual feature and each text prototype. And the zero-shot prediction $q_i \in \mathbb{R}^C$ is then obtained via a scaled softmax:
\begin{equation} 
s_{i,c} = \mathbf{z}_i^\top \mathbf{w}_c, \qquad q_{i,c} = \frac{\exp(\kappa \cdot s_{i,c})}{\sum_{j=1}^{C} \exp(\kappa \cdot s_{i,j})}, 
\label{eq:zeroshot} \end{equation}
where $\kappa$ is the predefined inverse temperature scalar. The maximum predictive confidence and the corresponding pseudo-label are denoted by $q_i^{\max} = \max_c q_{i,c}$ and $\hat{y}_i = \arg\max_c q_{i,c}$, respectively.
% At test time, we observe an unlabeled target stream of mini-batches $\{\mathcal{B}_t\}_{t=1}^{B}$, where
% $\mathcal{B}_t=\{\mathbf{x}_{1},\dots,\mathbf{x}_{t}\}$.
% For simplicity, we omit the stream index $t$ below and describe the adaptation on a generic test batch.
% For an input image $\mathbf{x}_i$, the visual encoder extracts a normalized feature
% $\mathbf{z}_i=f_{\theta}(\mathbf{x}_i)\in\mathbb{R}^d$.
% Let $\{\mathbf{w}_c\}_{c=1}^{C}$ denote the frozen normalized text prototypes.
% We  then define the cosine similarity score, scaled logit and zero-shot predictive distribution by
% \begin{equation}
% s_{i,c}=\mathbf{z}_i^\top \mathbf{w}_c,
% \qquad
% o_{i,c}=\kappa \cdot s_{i,c},
% \qquad
% q_{i,c}
% =
% \frac{\exp(o_{i,c})}{\sum_{j=1}^{C}\exp(o_{i,j})},
% \label{eq:zeroshot}
% \end{equation}
% where $\kappa$ is the predefined inverse temperature scalar from the CLIP model. The maximum predictive confidence and current top prediction are further given by
% \begin{equation}
% q_i^{\max}=\max_c q_{i,c},
% \qquad
% \hat{y}_i=\arg\max_c q_{i,c}.
% \end{equation}

\paragraph{Online Test-Time Adaptation.}
We consider a strict online test-time adaptation paradigm. At each time step $t$, the pre-trained VLM performs a single-step update on an unlabeled batch $\mathcal{B}_t$, without access to source data, replay buffers, or test labels.
To adapt the model, we update only the affine parameters (denoted by $\theta$) of the LayerNorm layers at the visual branch. All remaining visual parameters, alongside the text encoder $f_{\phi}$ and the prototypes $\{\mathbf{w}_c\}_{c=1}^{C}$, are kept frozen throughout the adaptation.

\subsection{Failure Regime and Design Principles}
\label{sec:failure_regimes}

% To motivate our design, we examine how CLIP predictions degrade under test-time shift. Figure~\ref{fig:obs_overview} summarizes the resulting failure patterns. Rather than viewing test-time failure merely as a uniform loss of confidence, we characterize it through two coupled levels of degradation that directly shape our method.
We analyze the failure patterns of test-time adaptation (TTA) for Vision-Language Models (VLMs) across two coupled dimensions: sample-level geometry and stream-level effects. 
% Our analysis reveals how the iterative suppression of locally recoverable ground truths leads to the accumulation of global prediction biases under distribution shifts.

\noindent\textbf{Sample-Level Geometry}. Standard TTA relies on entropy minimization, assuming the top-1 prediction is reliable for adaptation. However, distribution shifts frequently corrupt this assumption. To locate the potential ground truth during these failures, we examine the local geometric structure of predictions via an oracle analysis on the corrupted data. For a misclassified sample $i$, the similarity gap between the incorrect prediction $\hat{y}_i$ and the ground-truth $y_i$ is defined as:
\begin{equation}
\Delta_i = s_{i,\hat{y}_i} - s_{i,y_i},
\end{equation}
As shown in Figure \ref{fig:obs_overview}(a), a substantial fraction of misclassified samples maintain a narrow gap ($\Delta_i \le \delta$). This proximity highlights the structural flaw of entropy minimization: sharpening an incorrect top-1 prediction actively suppresses the nearby potential ground truth. Given the absence of ground-truth labels during testing, a robust strategy should preserve the entire near-top candidate set rather than committing to a single class.

% We observe that a substantial fraction of misclassified samples maintain a narrow gap ($\Delta_i \le \delta$), as shown by the shaded region in Figure \ref{fig:obs_overview}(a). This observation reveals that even during top-1 failures, corrupted visual features remain in close proximity to their correct semantic anchors. This proximity explains why entropy minimization becomes counterproductive under shift: sharpening an incorrect top prediction suppresses the nearby ground truth. This directly motivates an alternative approach that preserves the near-top candidate set.

Let $\mathcal{P}_i=\{c \mid s_{i,c}\ge s_{i,\hat{y}_i}-\delta\}$ denote this plausible near-top candidate set. However, the internal geometry of this set remains highly vulnerable. To quantify this, we measure the local margin between the weakest candidate in $\mathcal{P}_i$ and the strongest outside competitor (hard negative):
\begin{equation}
m_i^{\mathrm{local}} = \min_{c\in\mathcal{P}i(\delta)} s_{i,c} - \max_{c\notin\mathcal{P}i(\delta)} s_{i,c}.
\end{equation}
Mathematically, a vanishing margin indicates that external hard negatives begin to outscore the candidates within $\mathcal{P}_i$. This score overlap compromises the geometric structure of the candidate set, leaving the potential ground truth vulnerable to severe suppression. Figure \ref{fig:obs_overview}(b) confirms this degradation, revealing that increasing corruption severity progressively compresses this local margin. The above observation and analysis motivate our first design principle.

\begin{tcolorbox}[colback=gray!5!white, colframe=gray!75!black, arc=2pt, boxrule=0.5pt, left=4pt, right=4pt, top=4pt, bottom=4pt]
\textbf{\textit{Design Principle 1:}} Rather than forcing overconfident commitment to the current top-1 prediction, sample-level adaptation should explicitly preserve the plausible near-top candidates and restore their margin against hard negatives.
\end{tcolorbox}

\noindent\textbf{Stream-Level Effects.}
While Principle 1 effectively protects the potential ground truth at the sample level, its long-term effectiveness is compromised by a temporal vulnerability across the continuous data stream. Vision-Language Models (VLMs) exhibit initial output biases that introduce upward shifts in the similarity scores of specific classes. Because membership in $\mathcal{P}_i$ relies on these scores, classes subject to these shifts frequently satisfy the selection threshold across the data stream. Consequently, they are disproportionately included in the local candidate sets and repeatedly undergo margin restoration. This creates a feedback loop: continuously restoring local margins for these classes amplifies their global accumulation, eventually overriding the intended geometric improvements.

To quantify this temporal degradation, we track the class-wise prediction frequency across the adaptation stream. As shown in Figure \ref{fig:obs_overview}(c), the model's output distribution exhibits increasing skewness over the data stream. This progressive bias accumulation ultimately results in total class collapse, as confirmed in Figure \ref{fig:obs_overview}(d). These observations yield a critical clue: a static, uniform margin strategy is fundamentally flawed for long-term TTA. To prevent dominant classes from hijacking the adaptation process, each prediction requires a separate, dynamically regulated margin calibrated by its global accumulation history. This dynamic requirement directly motivates our second design principle.

% this output preference can further distort the online adaptation process. At adaptation stage, each update is still anchored on the current top-1 image--text match. As a result, pseudo-classes that are already slightly preferred in the emitted predictions are also more likely to trigger subsequent updates and receive more restoration pressure. If adaptation applies only local restoration on a batch-by-batch basis, without a stream-level mechanism to regulate how updates are allocated across pseudo-classes, this initial preference can be repeatedly fed back into the update process and gradually amplified.

% If adaptation relies solely on local sample-wise corrections without global constraints, these initial biases will be continuously reinforced throughout the online updating process.

% To quantify this temporal degradation, we track the class-wise prediction frequency across the data stream. As shown in Figure~\ref{fig:obs_overview}(c), he model's output increasingly deviates from a balanced distribution. These biased predictions serve as self-training signals, progressively shifting the model parameters toward a subset of dominant classes. As shown in Figure \ref{fig:obs_overview}(d), this process eventually leads to total class collapse, confirming that without global stabilization, local geometric improvements are overridden by stream-level bias accumulation.

\begin{tcolorbox}[colback=gray!5!white, colframe=gray!75!black, arc=2pt, boxrule=0.5pt, left=4pt, right=4pt, top=4pt, bottom=4pt]
\textbf{\textit{Design Principle 2:}} For long-term effectiveness of Principle 1, preventing progressive bias accumulation by establishing dynamically calibrated margins for each class is essential.
\end{tcolorbox}

\begin{figure*}[t]
    \centering
    \includegraphics[width=\textwidth]{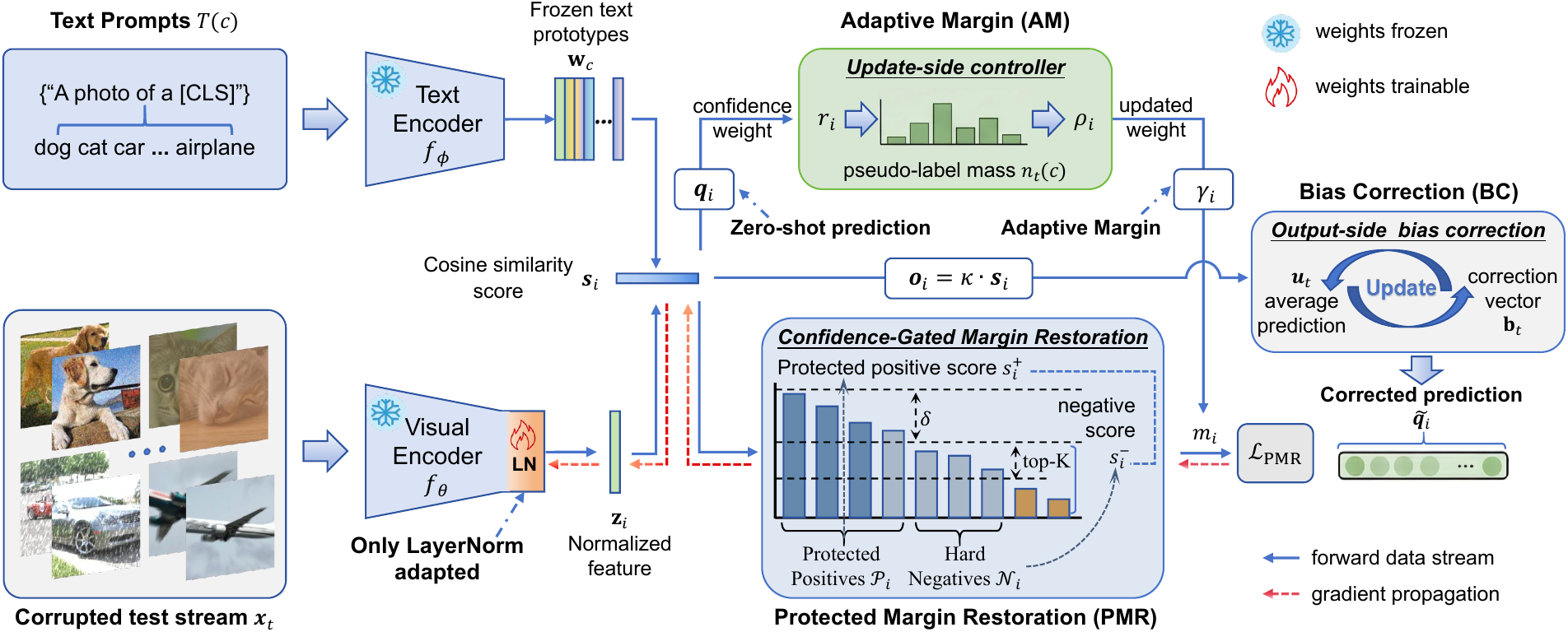}
    % \caption{Illustration of the proposed Local Margin Restoration (LMR) framework for online test-time adaptation of vision-language models. The proposed approach consists of three main components: (1) Protected Margin Restoration (PMR) for instance-level adaptation, (2) an update-side Adaptive Margin (AM) controller, and (3) an output-side Bias Correction (BC) module for stream-level stabilization.}
    \vspace{-0.5cm}
    \caption{\footnotesize{Illustration of the proposed Local Margin Restoration (LMR) framework for online test-time adaptation of vision-language models. The framework is composed of three core components: (1) Protected Margin Restoration (PMR) protects plausible near-top classes from premature suppression to restore local semantic geometry; (2) Adaptive Margin (AM) dynamically regulates the update-side margin based on pseudo-label accumulation; and (3) Bias Correction (BC) calibrates the output predictions to prevent long-horizon performance collapse.}}
    \Description{Architecture of the LMR framework showing the Protected Margin Restoration, Adaptive Margin, and Bias Correction modules.}
    \label{fig:framework}
\end{figure*}

Based on the above observations, we design an online  test-time adaptation framework that combines sample-level protected margin restoration with stream-level stabilization. Figure~\ref{fig:framework} provides an overview of our method.

\subsection{Protected Set-Anchored Margin Restoration}
\label{sec:pmr}
Our core adaptation objective is \emph{Protected Candidate-Set Margin Restoration} (PMR), which directly targets the recoverable local-neighborhood regime described above. Rather than sharpening the current top-1 prediction, PMR explicitly constructs a protected candidate set of plausible near-top classes on the positive side and restores their local semantic margin against hard competitors outside that set.

\noindent\textbf{Protected Candidate Set.}
Following the analysis in Sec.~\ref{sec:failure_regimes}, we construct a protected candidate set around the current top prediction:
\begin{equation}
\mathcal{P}_i
=
\left\{
c \in \mathcal{Y}
\;\middle|\;
s_{i,c} \ge s_{i,\hat{y}_i} - \delta
\right\},
\label{eq:candidate_set}
\end{equation}
where $\delta>0$ is a score-gap threshold. The set $\mathcal{P}_i$ contains classes whose scores remain sufficiently close to the current top prediction, and these classes are therefore protected from being immediately treated as negatives during adaptation.

\noindent\textbf{Soft Positive Aggregation.}
Within the candidate set, we aggregate the positive logits using normalized zero-shot posterior weights:
\begin{equation}
\alpha_{ic}
=
\frac{
q_{ic} \cdot \mathbb{I}[c \in \mathcal{P}_i]
}{
\sum_{j=1}^{C} q_{ij} \cdot \mathbb{I}[j \in \mathcal{P}_i] + \varepsilon
},
\label{eq:candidate_weights}
\end{equation}
where $\varepsilon$ is a small constant for numerical stability. The protected positive score is then defined as
\begin{equation}
s_i^{+}
=
\log
\sum_{c \in \mathcal{P}_i}
\alpha_{i,c}\exp(o_{i,c}),
% =
% \log
% \sum_{c \in \mathcal{P}_i}
% \exp\!\left(o_{i,c}+\log \alpha_{i,c}\right).
\label{eq:positive_score}
\end{equation}
where $o_{i,c}=\kappa\cdot s_{i,c}$ indicates the scaled cosine similarity score.
% This design keeps all candidate classes on the positive side, while allowing the supervision to remain soft within the protected neighborhood.

\noindent\textbf{Hard Negative Separation.}
To restore the local separation, we do not contrast the protected set against all remaining classes equally. Instead, we focus on the hardest external competitors outside the candidate set. Specifically, we define $\mathcal{N}_i
=
\operatorname*{arg\,topK}_{c \notin \mathcal{P}_i} \; s_{i,c}$,
% \begin{equation}
% \mathcal{N}_i = \operatorname{top-K}\!\left( \{\ell_{ic} \mid c \notin \mathcal{P}_i\}, K \right),
% \label{eq:negative_set}
% \end{equation}
where $K$ is the number of hard negatives. 
The corresponding negative score is
\begin{equation}
s_i^{-}
=
\log\!\left(
\frac{1}{K}
\sum_{c \in \mathcal{N}_i}
\exp(o_{ic})
\right).
\label{eq:negative_score}
\end{equation}
By restricting negative pressure to hard competitors outside the protected neighborhood, PMR avoids explicitly pushing away still-plausible nearby classes.

\noindent\textbf{Confidence-Gated Margin Restoration.}
Using Eqs.~\eqref{eq:positive_score} and \eqref{eq:negative_score}, we define the protected local margin of sample $i$ as $m_i = s_i^{+} - s_i^{-}$
% \begin{equation}
% m_i = s_i^{+} - s_i^{-},
% \label{eq:pmr_margin}
% \end{equation}
and restore this margin toward a target value $\gamma>0$ via
\begin{equation}
\mathcal{L}_i^{\mathrm{PMR}}
=
\operatorname{softplus}(\gamma - m_i).
\label{eq:pmr_loss}
\end{equation}

To suppress unreliable updates, we weight each sample by
\begin{equation}
    r_i
=
\left(
\frac{q_i^{\max}-\tau_{\mathrm{conf}}}{1-\tau_{\mathrm{conf}}}
\right)
\cdot \mathbb{I}[q_i^{\max}\ge\tau_{\mathrm{conf}}],
\label{eq:r_i}
\end{equation}
where $q_i^{\max}$ is the maximum posterior probability and $\tau_{\mathrm{conf}}$ is a fixed confidence threshold.
The batch PMR objective is
\begin{equation}
\mathcal{L}_{\mathrm{PMR}}
=
\frac{\sum_i r_i\,\mathcal{L}_i^{\mathrm{PMR}}}
{\sum_j r_j + \epsilon},
\label{eq:pmr_batch_loss}
\end{equation}
% which focuses margin restoration on confident and recoverable samples.

% \paragraph{Discussion.}
% PMR is not a generic soft pseudo-labeling variant. Standard full-distribution soft matching does not explicitly distinguish \emph{recoverable near-top classes} from \emph{hard external competitors}, nor does it directly optimize the local margin between them. The asymmetry in PMR is deliberate: the positive side is softened only within the detached self-anchored candidate set, because the goal there is to preserve still-plausible nearby classes instead of prematurely collapsing them to a single winner; the negative side is hardened outside the set, because the goal is to restore separation from the most competitive external distractors rather than to average over many weak or irrelevant classes. A hard top-1 positive would reintroduce \emph{recoverable-class exclusion}, whereas diffuse soft negatives would dilute the boundary-restoration signal. Therefore, PMR is a \emph{set-structured local margin restoration} objective: it protects plausible nearby positives, confines negative pressure to external hard competitors, and directly targets the failure regime in Sec.\ref{failue_regime}. 

\subsection{Stream-Level Stabilization}
\label{sec:stream_stabilization}

While PMR addresses the sample-level geometry in Principle~1, stable online adaptation also requires controlling how restoration pressure accumulates over the test stream. As discussed in Sec.~\ref{sec:failure_regimes}, a static and uniform margin is vulnerable to progressive bias accumulation: classes that are repeatedly favored by the current model tend to receive restoration more often, and such repeated reinforcement can gradually dominate the stream. To implement Principle~2 directly, we therefore replace the static PMR margin with a dynamically calibrated one. In addition, we further introduce an output-side bias correction module as a complementary stabilizer. PMR remains the core sample-level restoration objective, while these two components improve its long-term stability.

\noindent\textbf{Update-Side Adaptive Margin.}
Although PMR preserves the plausible near-top set for each sample, restoration is still anchored on the current pseudo-class $\hat y_i$. Consequently, classes that are repeatedly favored under shift are also more likely to trigger restoration across the stream and absorb disproportionately large update mass. Over time, this self-reinforcing process makes a static margin increasingly inappropriate, since it allocates similar restoration strength to pseudo-classes with very different accumulation histories.

To quantify this effect, we maintain a running pseudo-class mass
\begin{equation}
n_t(c)
=
n_{t-1}(c)
+
\sum_{i\in\mathcal{B}_t}
r_i\,\mathbb{I}[\hat y_i=c],
\label{eq:nt_update_final}
\end{equation}
where $r_i$ is the confidence-based sample weight defined in Eq.~\eqref{eq:r_i}. This quantity summarizes how much reliable restoration mass has already been assigned to pseudo-class $c$ over the stream. Although output-side preference may first manifest itself through more frequent inclusion in the protected set, update-side accumulation is ultimately driven by which pseudo-class repeatedly serves as the restoration anchor. We therefore use hard anchor occupancy as the stream-level signal for update-side reinforcement.

For sample $i$, we define the running accumulation share of its current pseudo-class as
\begin{equation}
\rho_i
=
\frac{n_t(\hat y_i)}
{\sum_{c'} n_t(c')+\varepsilon}.
\label{eq:rho_i_final}
\end{equation}
In implementation, we instantiate $\rho_i$ with a leave-one-out readout of the confidence-weighted hard pseudo-class counts in Eq.~\eqref{eq:nt_update_final}, which avoids immediate self-reinforcement within the same batch.

We then replace the static PMR target margin with a sample-specific one:
\begin{equation}
\gamma_i
=
\tilde{\gamma}_0
-
\beta \log(\rho_i+\varepsilon),
\label{eq:adaptive_gamma_final}
\end{equation}
where $\tilde{\gamma}_0$ is the absorbed base margin and $\beta>0$ controls the feedback strength. Pseudo-classes that have already absorbed more updated mass receive a smaller target margin and hence weaker further reinforcement, whereas under-represented pseudo-classes retain stronger restoration pressure. In this way, adaptive margin implements Principle~2 directly: it regulates restoration strength according to stream history while leaving the locally plausible candidate set itself unchanged.

\noindent\textbf{Output-Side Residual Bias Correction.}
While adaptive margin mitigates update-side self-reinforcement, it does not directly correct persistent class-wise preference in the emitted logits. Under test-time shift, some classes can remain systematically over-scored even before online updates accumulate, and this effect may persist even when update-side concentration is controlled. Since PMR and adaptive margin regulate how restoration pressure is allocated during adaptation rather than the logits themselves, they do not explicitly remove such residual output-side bias.

To compensate for this effect, we introduce a lightweight closed-loop controller in logit space. Specifically, we maintain an additive correction vector $\mathbf{b}_t \in \mathbb{R}^C$ and apply the previous correction to the current logits:
\begin{equation}
\tilde{\mathbf{o}}_i = \mathbf{o}_i + \mathbf{b}_{t-1},
\label{eq:corrected_logits}
\end{equation}
where $\mathbf{o}_i$ denotes the uncorrected logits and $\tilde{\mathbf{o}}_i$ the corrected ones. The controller is initialized with $\mathbf{b}_0=\mathbf{0}$.

A single batch is too noisy to reveal persistent stream-level preference, so we estimate class-wise output concentration through a running average prediction:
\begin{equation}
\mathbf{u}_t
=
\frac{1}{|\mathcal{B}_t|}
\sum_{i\in\mathcal{B}_t}
\operatorname{softmax}(\tilde{\mathbf{o}}_i),
\qquad
\bar{\mathbf{u}}_t
=
\mu \bar{\mathbf{u}}_{t-1}
+
(1-\mu)\mathbf{u}_t,
\label{eq:running_pred_avg}
\end{equation}
where $\mu \in [0,1)$ is the EMA momentum. In practice, we bootstrap the controller by setting $\bar{\mathbf{u}}_1=\mathbf{u}_1$.

An additive correction in logit space corresponds to a multiplicative reweighting in probability space. Accordingly, if a class persistently absorbs excessive predictive mass over the stream, decreasing its logit by an amount proportional to the log of its running average prediction provides a natural counterbalance. Since softmax is invariant to a global shift, only relative class offsets need to be corrected. We therefore update
\begin{equation}
\mathbf{b}_t
=
\mathbf{P}\!\left(
\mathbf{b}_{t-1}
-
\eta \log(\bar{\mathbf{u}}_t+\varepsilon)
\right),
\qquad
\mathbf{P}
=
\mathbf{I}
-
\frac{1}{C}\mathbf{e}\mathbf{e}^{\top},
\label{eq:bias_update}
\end{equation}
where $\eta>0$ is the feedback gain, $\varepsilon$ is a small constant for numerical stability, $\mathbf{e}\in\mathbb{R}^{C}$ is the all-ones vector, and $\mathbf{P}$ projects onto the zero-mean subspace. The updated $\mathbf{b}_t$ is then used at the next step, yielding a simple feedback loop that gradually suppresses persistent over-scored classes.

This module complements adaptive margin at a different level: adaptive margin regulates how restoration pressure accumulates through repeated updates, whereas residual bias correction directly compensates persistent class-wise preference in the emitted logits.

\subsection{Gradient Insights}
\label{sec:theoretical_insights}

We provide a gradient-based interpretation of LMR and the stream-level stabilizers. Our goal here is not to establish a global convergence theorem for the full online adaptation process, but to clarify the effective update structure induced by each component. Detailed derivations are deferred to the appendix. Throughout this subsection, we condition on the detached protected set $\mathcal{P}_i$, detached candidate weights $\alpha_{ic}$, the detached hard-negative set $\mathcal{N}_i$, and the scalar target margin $\gamma_i$, and consider points away from support-switching events induced by top-$K$ selection. For clarity, we omit the detached sample weight and the common batch-averaging constant, since they only rescale the contribution of this sample and do not change its update direction.

\noindent\textbf{Protected gradient structure.}
Recall that LMR optimizes the protected margin
\(
m_i=s_i^{+}-s_i^{-}
\)
through
\(
\mathcal{L}_i^{\mathrm{LMR}}=\operatorname{softplus}(\gamma_i-m_i).
\)
Let
\(
\nu_i=\sigma(\gamma_i-m_i).
\)
, where $\sigma$ is the sigmoid function. Define the normalized positive and negative weights as
\begin{equation}
\pi_{ic}^{+}
=
\frac{
\alpha_{ic}\exp(o_{ic})
}{
\sum_{j\in\mathcal{P}_i} \alpha_{ij}\exp(o_{ij})
},
\qquad
\pi_{ic}^{-}
=
\frac{
\exp(o_{ic})
}{
\sum_{j\in\mathcal{N}_i}\exp(o_{ij})
}.
\label{eq:lmr_pi}
\end{equation}
Then the per-sample logit gradient takes the following piecewise form:
\begin{equation}
\frac{\partial \mathcal{L}_i^{\mathrm{LMR}}}{\partial o_{ic}}
=
\begin{cases}
-\nu_i \pi_{ic}^{+}, & c\in\mathcal{P}_i,\\[3pt]
\phantom{-}\nu_i \pi_{ic}^{-}, & c\in\mathcal{N}_i,\\[3pt]
0, & c\notin \mathcal{P}_i\cup\mathcal{N}_i.
\end{cases}
\label{eq:lmr_logit_grad}
\end{equation}
This shows that LMR has an explicitly support-restricted gradient: classes inside the protected set are increased, masked external hard negatives are suppressed, and all remaining classes receive zero gradient. In particular, if the ground-truth class remains inside $\mathcal{P}_i$, it is not assigned explicit negative pressure. This explains why LMR is not equivalent to full-distribution soft pseudo-labeling: it restores a \emph{local} score gap around a protected near-top region rather than matching the full predictive distribution. When the protected set degenerates to a singleton, i.e., $\mathcal{P}_i=\{\hat y_i\}$, only the current top-1 class receives positive-side support; recoverable near-top classes outside $\mathcal{P}_i$ lose that protection, which explains why the top-1-only variant is weaker when multiple classes remain close in score.

% \noindent\textbf{Contrast to entropy minimization.}
% For comparison, entropy minimization induces a dense logit gradient over all classes, since
% \begin{equation}
% \mathcal{L}_i^{\mathrm{ent}}
% =
% -\sum_c q_{ic}\log q_{ic}
% \quad\Longrightarrow\quad
% \frac{\partial \mathcal{L}_i^{\mathrm{ent}}}{\partial o_{ic}}
% =
% q_{ic}\bigl(-\mathcal{H}(\mathbf q_i)-\log q_{ic}\bigr),
% \label{eq:entropy_grad}
% \end{equation}
% where $\mathcal{H}(\mathbf q_i)$ is the entropy of the predictive distribution. Thus, entropy minimization globally sharpens the full output distribution and can reinforce an incorrect top-1 prediction. In contrast, LMR assigns zero gradient outside $\mathcal{P}_i\cup\mathcal{N}_i$ and therefore focuses the update on restoring a local near-top margin.

\noindent\textbf{Contrast to entropy minimization.}
Entropy minimization globally sharpens the predictive distribution and can therefore reinforce an incorrect top-1 prediction. Specifically, for any class $c$ with $q_{i,c}>0$, the entropy objective induces a non-zero logit gradient,
\begin{equation*}
\frac{\partial \mathcal L_i^{\mathrm{ent}}}{\partial o_{i,c}}
=
q_{i,c}\big(-\mathcal H(\mathbf q_i)-\log q_{i,c}\big),
\end{equation*}
which is dense over the full output space (detailed derivations can be found in the appendix). In effect, the gradient drives the prediction toward lower entropy by increasing high-probability classes and suppressing low-probability ones, regardless of whether they are locally relevant to correcting the current mistake. By contrast, our objective assigns zero gradient outside $\mathcal{P}_i\cup\mathcal{N}_i$, thereby confining the update to restoring the local score gap among plausible candidates rather than globally sharpening the whole distribution.

% \medskip
% \noindent\textbf{Representation-level pressure.}
% Since the logits satisfy \(o_{ic}=\kappa\,\mathbf{z}_i^{\top}\mathbf{w}_c\), Eq.~\eqref{eq:lmr_logit_grad} induces the following effective representation-level pressure:
% \begin{equation}
% -\nabla_{\mathbf{z}_i}\mathcal{L}_i^{\mathrm{LMR}}
% =
% \kappa\nu_i
% \left(
% \sum_{c\in\mathcal{P}_i}\pi_{ic}^{+}\mathbf{w}_c
% -
% \sum_{c\in\mathcal{N}_i}\pi_{ic}^{-}\mathbf{w}_c
% \right).
% \label{eq:lmr_feature_grad}
% \end{equation}
% Equation~\eqref{eq:lmr_feature_grad} characterizes the desired pressure that LMR exerts on the representation through the logits: it pulls the feature toward a weighted average of the protected candidate prototypes while pushing it away from external hard competitors. The actual parameter update follows the chain rule through the visual encoder and its LayerNorm parameters, but this logit-level analysis captures the intended direction of representation change.

\noindent\textbf{Adaptive margin as negative feedback.}
Under the same conditioning, the adaptive margin does not change the support pattern in Eq.~\eqref{eq:lmr_logit_grad}; it only rescales the update amplitude through $\nu_i=\sigma(\gamma_i-m_i)$. Substituting the adaptive margin definition from Eq.~\eqref{eq:adaptive_gamma_final} yields
\begin{equation}
\frac{\partial \nu_i}{\partial \rho_i}
=
-\frac{\beta}{\rho_i+\epsilon}\,\nu_i(1-\nu_i) < 0.
\label{eq:adaptive_margin_feedback}
\end{equation}
Hence, the restoration pressure decreases monotonically as the running pseudo-class mass grows. This establishes the adaptive margin as a negative-feedback controller that counteracts repeated reinforcement of already dominant pseudo-classes in the stream.

\section{Experiments}

\begin{table*}[!t]
\centering
\caption{{Classification accuracy (\%) on CIFAR-10-C, CIFAR-100-C, and ImageNet-C across 15 corruption types at severity level 5. All evaluated methods are based on the ViT-B/16 backbone. We compare our approach with state-of-the-art Zero-Shot and Test-Time Adaptation (TTA) methods. \textbf{Bold} and \underline{underline} denote the best and second-best results, respectively.}\vspace{-0.3cm}}
\label{tab:main_results}
% \small
\setlength{\tabcolsep}{3.2pt} % 优化列间距
\resizebox{0.9\textwidth}{!}{
% \begin{tabular}{l|ccc|cccc|ccc|ccccc|c}
\begin{tabular}{l|ccccccccccccccc|c}
\toprule
\textbf{Method} & \textbf{Gauss.} & \textbf{Shot} & \textbf{Impu.} & \textbf{Defo.} & \textbf{Glas.} & \textbf{Moti.} & \textbf{Zoom} & \textbf{Snow} & \textbf{Fros.} & \textbf{Fog} & \textbf{Brit.} & \textbf{Cont.} & \textbf{Elas.} & \textbf{Pixl.} & \textbf{Jpeg} & \textbf{AVG} \\
% \multirow{2}{*}{\textbf{Method}} & \multicolumn{3}{c|}{Noise} & \multicolumn{4}{c|}{Blur} & \multicolumn{3}{c|}{Weather} & \multicolumn{5}{c|}{Digital} & \multirow{2}{*}{\textbf{AVG}} \\
% \cmidrule(lr){2-4} \cmidrule(lr){5-8} \cmidrule(lr){9-11} \cmidrule(lr){12-16}
%  & Gauss. & Shot & Impu. & Defo. & Glas. & Moti. & Zoom & Snow & Fros. & Fog & Brit. & Cont. & Elas. & Pixl. & Jpeg & \\
\midrule
\multicolumn{17}{c}{\textit{CIFAR-10-C}} \\
\midrule
CLIPZS~\cite{radford21clip}  & 37.91 & 41.66 & 54.42 & 71.75 & 40.89 & 67.86 & 73.61 & 73.84 & 77.34 & 70.26 & 84.42 & 62.32 & 53.82 & 47.60 & 59.44 & 61.14 \\
TDA~\cite{karmanov2024efficient}     & 41.38 & 45.30 & 57.55 & 73.45 & 46.33 & 70.57 & 75.86 & 75.62 & 79.20 & 71.19 & 85.66 & 64.10 & 56.20 & 52.45 & 60.38 & 63.68 \\
ECALP~\cite{li2025efficient}   & 45.00 & 49.13 & 60.03 & 75.41 & 44.90 & 71.82 & 77.38 & 76.74 & 79.54 & 71.93 & 86.62 & 65.25 & 58.53 & 52.80 & 61.95 & 65.14 \\
TENT~\cite{wang2021tent}    & 16.22 & 19.31 & 44.74 & 80.19 & 24.53 & 76.19 & 82.05 & \underline{83.35} & 83.31 & {81.01} & \underline{91.25} & 81.85 & 62.71 & \underline{67.66} & 57.74 & 63.47 \\
ETA~\cite{niu22Efficient}     & 39.15 & 42.81 & 54.49 & 71.67 & 41.44 & 67.94 & 73.80 & 73.91 & 77.40 & 70.30 & 84.48 & 62.35 & 53.93 & 47.86 & 59.52 & 61.40 \\
BATCLIP~\cite{maharana2025batclip} & {61.08} & 64.07 & 65.46 & \underline{80.47} & {55.02} & \underline{80.71} & \underline{81.91} & 83.04 & \underline{84.20} & 80.90 & 88.92 & \underline{82.18} & {69.16} & 62.77 & {67.56} & {73.83} \\
MINT~\cite{baomint}    & 60.44 & {64.36} & \underline{69.57} & 78.70 & 52.70 & 80.43 & 80.57 & 81.72 & 82.23 & 79.89 & 89.94 & 79.13 & 65.31 & 62.82 & 66.83 & 72.98 \\
BITTA~\cite{sun2026bilateral} & \underline{61.92} & \underline{66.43} & 64.97 & 79.63 & \underline{58.14} &  80.02 & 80.61 & 81.72 & 83.71 & \underline{81.10} & 86.76 & 81.69 & \underline{70.89} & 64.91 & \underline{68.38} & \underline{74.06} \\
\midrule
\textbf{Ours} & \textbf{66.83} & \textbf{67.98} & \textbf{72.22} & \textbf{82.83} & \textbf{64.01} & \textbf{82.81} & \textbf{83.39} & \textbf{85.06} & \textbf{86.06} & \textbf{84.71} & \textbf{91.92} & \textbf{86.15} & \textbf{73.23} & \textbf{71.51} & \textbf{69.94} & \textbf{77.91} \\
\midrule
\multicolumn{17}{c}{\textit{CIFAR-100-C}} \\
\midrule
CLIPZS~\cite{radford21clip}  & 19.64 & 21.45 & 25.37 & 42.50 & 20.10 & 43.20 & 48.00 & 49.73 & 41.64 & 45.00 & 57.00 & 34.45 & 29.19 & 23.94 & 32.49 & 35.80 \\
TDA~\cite{karmanov2024efficient}     & 22.42 & 24.46 & 31.68 & 44.57 & 21.29 & 44.59 & 49.81 & 48.95 & 51.59 & 43.16 & 58.86 & 35.15 & 30.06 & 27.29 & 33.42 & 37.82 \\
ECALP~\cite{li2025efficient}   & 23.63 & 25.44 & 31.55 & 46.19 & 22.75 & 45.13 & 51.41 & 50.64 & 52.40 & 44.36 & 60.67 & 37.12 & 31.61 & 27.42 & 34.34 & 38.98 \\
TENT~\cite{wang2021tent}    & 8.01  & 8.82  & 9.26  & 51.46 & 8.35  & \underline{52.26} & \underline{55.42} & \underline{54.45} & 45.52 & \underline{50.78} & 65.48 & \underline{52.21} & \underline{36.32} & \underline{41.92} & \underline{38.90} & 38.61 \\
ETA~\cite{niu22Efficient}     & 29.57 & 31.77 & 35.25 & 47.47 & \underline{28.01} & 48.18 & 53.50 & 52.43 & \underline{53.01} & 50.42 & 62.75 & 49.11 & 33.70 & 39.12 & 38.27 & 43.50 \\
BATCLIP~\cite{maharana2025batclip} & 24.85 & 27.80 & 33.64 & 50.01 & 26.32 & 48.52 & 54.84 & 52.35 & 51.68 & 48.37 & 63.29 & 45.12 & 34.88 & 32.43 & 37.31 & 42.09 \\
MINT~\cite{baomint}    & \underline{29.68} & \underline{33.72} & \underline{40.64} & \underline{50.89} & 27.66 & 49.38 & 55.33 & 52.49 & 52.20 & 49.94 & \underline{65.71} & 48.42 & 36.28 & 33.17 & 38.27 & \underline{44.25} \\
BITTA~\cite{sun2026bilateral} & 26.60 & 29.11 & 35.87 & 50.03 & 27.07 & 49.26 & 54.87 & 51.92 & 52.11 & 48.6 & 63.00 & 46.85 & 35.18 & 32.56 & 37.51 & 42.70 \\
\midrule
\textbf{Ours} & \textbf{38.91} & \textbf{41.44} & \textbf{48.47} & \textbf{56.79} & \textbf{38.61} & \textbf{57.15} & \textbf{61.36} & \textbf{60.70} & \textbf{61.07} & \textbf{59.00} & \textbf{70.36} & \textbf{63.17} & \textbf{45.35} & \textbf{51.58} & \textbf{45.39} & \textbf{53.29} \\
\midrule
\multicolumn{17}{c}{\textit{ImageNet-C}} \\
\midrule
CLIPZS~\cite{radford21clip}  & 11.20 & 12.48 & 12.04 & 23.36 & 15.16 & 24.48 & 22.68 & 32.36 & 29.82 & 35.82 & 54.02 & 17.30 & 12.76 & 31.04 & 33.34 & 24.52 \\
TDA~\cite{karmanov2024efficient}     & 12.38 & 13.72 & 12.82 & 23.30 & 15.38 & 24.98 & 22.64 & 32.54 & 31.10 & 36.50 & 54.36 & 17.54 & 13.70 & 32.08 & 33.70 & 25.12 \\
ECALP~\cite{li2025efficient}   & 13.98 & 15.78 & 14.52 & 25.58 & 16.96 & 26.16 & 25.28 & 34.40 & 32.94 & 39.06 & 56.00 & 19.92 & 15.54 & 34.48 & 35.56 & 27.08 \\
TENT~\cite{wang2021tent}    & 5.14  & 5.70  & 7.44  & 25.22 & 19.34 & 26.80 & 24.16  & 33.56 & 30.40 & 37.74 & 54.22 & 22.50 & 13.90 & 35.02  & 36.08 & 25.15 \\
ETA~\cite{niu22Efficient}     & 19.08 & 20.06 & 20.48 & 25.96 & \underline{22.86} & 29.58 & 26.80 & 35.30 & \underline{32.48} & 39.72 & 55.74 & {26.64} & 21.64 & 37.26 & 38.00 & 30.11 \\
BATCLIP~\cite{maharana2025batclip} & 19.18 & 21.58 & 19.68 & 26.72 & 21.92 & 30.84 & \underline{29.14} & \underline{36.54} & 31.94 & \underline{40.92} & \underline{56.64} & 26.18 & \underline{23.68} & 37.54 & 38.50 & \underline{30.73} \\
MINT~\cite{baomint}    & {20.26}	& {21.84}	& \underline{21.54}	& \underline{27.26}	& 22.26	& \underline{31.36}	& 27.70	& 34.04	& 30.64	& 40.88	& 55.02	& 26.06	& 20.58	& \underline{37.88}	& \underline{38.88}	& 30.41 \\
BITTA~\cite{sun2026bilateral} & \underline{20.76} & \underline{22.46} & 20.72 & 26.50 & 22.22 & 30.02 & 28.98 & 35.48 & 31.14 & 39.68 & 53.72 & \underline{27.24} & 22.26 & 35.94 & 36.58 & 30.25 \\
\midrule
\textbf{Ours} &\textbf{23.06}	&\textbf{25.46}	&\textbf{24.88}	&\textbf{30.22}	&\textbf{29.48}	&\textbf{35.98}	&\textbf{33.86}	&\textbf{41.60}	&\textbf{38.10}	&\textbf{46.22}	&\textbf{58.56}	&\textbf{34.10}	&\textbf{32.92}	&\textbf{43.66}	&\textbf{43.06}	&\textbf{36.08} \\
\bottomrule
\end{tabular}
}
\end{table*}

\subsection{Experimental Setup}
\noindent\textbf{Datasets and protocol.}
% We evaluate online test-time adaptation of CLIP under distribution shift on CIFAR-10-C~\cite{krizhevsky2009learning}, CIFAR-100-C\cite{krizhevsky2009learning}, and ImageNet-C~\cite{deng2009imagenet}, following the standard common-corruption benchmark~\cite{hendrycks2019robustness}. Unless otherwise specified, we report mean accuracy over all 15 corruption types at severity 5. Each corruption type is treated as an independent test stream, where unlabeled batches arrive sequentially and the model is updated once per batch without episodic reset. 
% In addition to common-corruption benchmarks, we also evaluate on ImageNet-A~\cite{hendrycks2021natural}, ImageNet-R~\cite{hendrycks2021many}, ImageNet-V2~\cite{recht2019imagenet} and ImageNet-S~\cite{wang2019learning} to assess robustness under broader distribution shifts, including adversarially filtered natural images, renditions, sketch-style shift, and a re-sampled test distribution. 
We evaluate online test-time adaptation of CLIP under distribution shift on CIFAR-10-C~\cite{krizhevsky2009learning}, CIFAR-100-C~\cite{krizhevsky2009learning}, and ImageNet-C~\cite{deng2009imagenet}, following the standard common corruption benchmark~\cite{hendrycks2019robustness}. Unless otherwise specified, we report mean accuracy over all 15 corruption types at severity 5, treating each corruption type as an independent unlabeled test stream with one update per batch and no episodic reset. We further evaluate on ImageNet-A~\cite{hendrycks2021natural}, ImageNet-R~\cite{hendrycks2021many}, ImageNet-V2~\cite{recht2019imagenet}, and ImageNet-Sketch~\cite{wang2019learning} to assess robustness under broader distribution shifts.

% \noindent\textbf{Implementation details.}
% We use ViT-B/16~\cite{dosovitskiy2020image} as the visual backbone of CLIP. Frozen text prototypes are constructed using the prompt template ``\textit{a photo of a class}.'' Our method updates only the affine parameters of the LayerNorm layers in the visual encoder, while keeping the text encoder fixed. We perform a single gradient step for each incoming batch. We use Adam~\cite{kingma2014adam} with learning rate 0.001 and batch size of 64. Unless otherwise specified, the key PMR hyperparameters are set to $\delta=0.03$, $\text{NegTopK}=5$ confidence threshold $\tau_{conf}$ of 0.3. The adaptive margin uses class-aware control, the base margin is $\tilde{\gamma}_0=\gamma_0-\beta\log C$, where coefficient $\beta=3.0$ for CIFAR10-C and 1.0 for others, $\gamma_0$ is the set to 0.3 for CIFAR10-C and 4.0 for others. The bias correction uses update rate $\eta=0.05$ with EMA momentum $m=0.8$. Additional implementation details are provided in the appendix.
\noindent\textbf{Implementation details.}
We use ViT-B/16~\cite{dosovitskiy2020image} as the visual backbone of CLIP. Frozen text prototypes are constructed using the prompt template ``\textit{a photo of a class}.'' Our method updates only the affine parameters of visual LayerNorm layers, while keeping the text encoder fixed. We perform one gradient step per incoming batch, using Adam~\cite{kingma2014adam} with learning rate 0.001 and batch size 64. Unless otherwise specified, the key PMR hyperparameters are $\delta=0.03$, $\text{NegTopK}=5$, and confidence threshold $\tau_{\mathrm{conf}}=0.3$. For adaptive margin, we use $\tilde{\gamma}_0=\gamma_0-\beta\log C$, with $\beta=3.0$ and $\gamma_0=0.3$ on CIFAR-10-C, and $\beta=1.0$ and $\gamma_0=4.0$ on the other benchmarks. Bias correction uses update rate $\eta=0.05$ and EMA momentum $m=0.8$. Additional details are provided in the appendix.

\noindent\textbf{Baselines.}
We compare our method against zero-shot CLIP~\cite{radford21clip} as the no-adaptation baseline and three groups of test-time adaptation baselines. The first group contains generic gradient-based TTA objectives adapted to CLIP, including TENT~\cite{wang2021tent} and ETA~\cite{niu22Efficient}. The second group contains CLIP-specific training-free adaptation methods, including TDA~\cite{karmanov2024efficient} and ECALP~\cite{li2025efficient}. The third group contains gradient-based adaptation methods specifically designed for CLIP, including BATCLIP~\cite{maharana2025batclip}, Mint~\cite{baomint}, and BITTA~\cite{sun2026bilateral}. Additional details of the baseline methods are provided in the appendix. 
% For methods whose original protocols differ from the standard online one-step setting, we adapt them conservatively to our evaluation protocol and summarize the corresponding changes in the appendix.

\subsection{Main Results}

\noindent\textbf{Results on Common Corruptions.}
Table~\ref{tab:main_results} reports results on CIFAR-10-C, CIFAR-100-C, and ImageNet-C. Our method achieves the best mean accuracy on all three benchmarks, reaching 77.91\%, 53.29\%, and 36.08\%, respectively. This corresponds to gains of +16.77, +17.49, and +11.56 points over zero-shot CLIP, and +3.85, +9.04, and +5.35 points over the strongest competing baseline on each dataset. Notably, our method ranks first on all 15 corruption types across all three datasets, showing that the gains are broad rather than corruption-specific. Compared with generic gradient-based TTA baselines, our method is substantially more robust under severe corruption; compared with CLIP-specific baselines, it remains consistently stronger across datasets, outperforming BATCLIP and BITTA on CIFAR-10-C and ImageNet-C, and Mint on CIFAR-100-C. These results indicate that the proposed method generalizes robustly across diverse corruption regimes.

% \noindent\textbf{Results on ImageNet variants.}
% Beyond ImageNet-C, which mainly evaluates robustness to synthetic common corruptions, we further evaluate on ImageNet-A, ImageNet-R, ImageNet-Sketch, and ImageNet-V2, which cover a broader range of distribution shifts. Table~\ref{tab:ood_results} shows that our method achieves the best average accuracy of 60.74\%, outperforming zero-shot CLIP by +3.59 points and the strongest competing baseline by +1.85 points. In particular, our method ranks first on ImageNet-A, ImageNet-R, and ImageNet-V2, improving over the strongest competing baseline by +1.19, +1.96, and +0.70 points, respectively. On ImageNet-Sketch, it achieves the second-best result and remains within only 0.19 points of the best method. Notably, the strongest competing baseline varies across these benchmarks, whereas our method remains consistently strong and delivers the best overall average. These results indicate that the proposed adaptation strategy generalizes beyond synthetic common corruptions and transfers effectively to broader ImageNet-style distribution shifts.
\noindent\textbf{Results on ImageNet variants.}
Beyond ImageNet-C, we further evaluate on ImageNet-A, ImageNet-R, ImageNet-Sketch, and ImageNet-V2, which cover broader distribution shifts. As shown in Table~\ref{tab:ood_results}, our method achieves the best average accuracy, improving over zero-shot CLIP by +3.59 points and over the strongest competing baseline by +1.85 points. It ranks first on ImageNet-A, ImageNet-R, and ImageNet-V2, and achieves the second-best result on ImageNet-Sketch with only a 0.19-point gap to the best method. These results show that the proposed adaptation strategy transfers effectively beyond synthetic common corruptions to broader diverse domain  shifts.

\begin{table}[!t]
\centering
\caption{{Classification accuracy (\%) on ImageNet-Variant benchmarks. \textbf{Bold} and \underline{underline} denote the best and second-best results, respectively.}\vspace{-0.3cm}}
\label{tab:ood_results}
\begin{tabular*}{0.9\linewidth}{@{\extracolsep{\fill}}lcccc|c}
\toprule
\textbf{Method} & \textbf{IN-A} & \textbf{IN-R} & \textbf{IN-S} & \textbf{IN-V2} & \textbf{AVG} \\
\midrule
CLIPZS  & 47.73 & 73.99 & 46.12 & 60.75 & 57.15 \\
TDA     & 49.28 & 75.47 & 48.88 & 61.20 & 58.71 \\
ECALP   & 47.37 & 75.66 & \textbf{50.70} & \underline{61.50} & 58.81 \\
TENT    & 48.31 & 75.49 & 46.88 & 60.91 & 57.90 \\
ETA     & 48.33 & \underline{75.92} & 48.29 & 60.97 & 58.38 \\
BATCLIP & 48.85 & 72.34 & 39.95 & 60.48 & 55.41 \\
MINT    & \underline{51.16} & 75.81 & 48.24 & 60.34 & \underline{58.89} \\
BITTA	& 48.43	& 70.00	& 36.97 & 55.70	& 52.78 \\
\midrule
\textbf{Ours} & \textbf{52.35} & \textbf{77.88} & \underline{50.51} & \textbf{62.20} & \textbf{60.74} \\
\bottomrule
\end{tabular*}
\vspace{-0.5cm}

\end{table}

\subsection{Ablation Study and Analysis}

\noindent\textbf{Ablation on components.}
Table~\ref{tab:component_ablation} reports the component ablation of our method. PMR alone is insufficient for stable stream-wise adaptation: relative to zero-shot CLIP, it improves ImageNet-C from 24.52 to 33.27, but brings only marginal gain on CIFAR-100-C (35.80 to 35.99) and even degrades CIFAR-10-C (61.14 to 58.23). This suggests that restoring local margins is beneficial, but its net effect depends on whether the update process remains stable over the stream. In contrast, adding adaptive margin to PMR yields substantial gains on all three datasets, improving PMR by +19.15, +15.08, and +1.08 on CIFAR-10-C, CIFAR-100-C, and ImageNet-C, respectively. This supports the view that repeated one-step updates otherwise over-concentrate on a small subset of pseudo-class anchors. Bias correction alone also improves over zero-shot CLIP, but it does not substitute for adaptive margin: PMR+BC remains consistently weaker than PMR+AM. This suggests that update-side concentration and output-side class-wise preference are related but distinct failure modes. Finally, combining adaptive margin and bias correction gives the best results on all datasets, indicating that the two components are complementary: adaptive margin stabilizes the update process, while bias correction further compensates residual class-wise preference in the output logits.

% \begin{table}[t]
% \centering
% \caption{{Component ablation on CIFAR10-C, CIFAR100-C, and ImageNet-C. AM denotes adaptive margin and BC denotes bias correction. We report mean accuracy (\%).}\vspace{-0.3cm}}
% \label{tab:component_ablation}
% \setlength{\tabcolsep}{4pt}
% \begin{tabular*}{0.9\linewidth}{@{\extracolsep{\fill}}lccccc@{}} % 修正为6列
% \toprule
% % 使用 \multicolumn 强制让表头这一格居中，同时不影响下面行左对齐
% \multirow{2}{*}{\makecell[c]{\textbf{Variants}}} & \multicolumn{2}{c}{\textbf{Components}} & \multicolumn{3}{c}{\textbf{Mean Accuracy (\%)}} \\
% \cmidrule(lr){2-3} \cmidrule(lr){4-6} % 修正跨度：Components占2列，Accuracy占3列
%  & \textbf{AM} & \textbf{BC} & \textbf{C10-C} & \textbf{C100-C} & \textbf{IN-C} \\
% \midrule
% ZS         & -      & -      & 61.14 & 35.80 & 24.52 \\
% Bias only  & -      & \checkmark & 67.06 & 40.28 & 28.07 \\
% PMR only   & -      & -      & 58.23 & 35.99 & 33.27 \\
% PMR + AM   & \checkmark & -      & 77.38 & 51.07 & 34.35 \\
% PMR + BC   & -      & \checkmark & 59.87 & 38.48 & 35.11 \\
% Full       & \checkmark & \checkmark & \textbf{77.91} & \textbf{53.29} & \textbf{36.08} \\
% \bottomrule
% \end{tabular*}
% \vspace{-0.3cm}
% \end{table}

\begin{table}[t]
\centering
\caption{Component ablation on CIFAR10-C, CIFAR100-C, and ImageNet-C.
AM denotes adaptive margin, and BC denotes bias correction.
We report mean accuracy (\%).}
\vspace{-0.3cm}
\label{tab:component_ablation}

\setlength{\tabcolsep}{4pt}
\begin{tabular*}{0.9\linewidth}
{@{\extracolsep{\fill}}lcccccc@{}}
\toprule
\multirow{2}{*}{\makecell[c]{\textbf{Variants}}}
& \multicolumn{3}{c}{\textbf{Components}}
& \multicolumn{3}{c}{\textbf{Mean Accuracy (\%)}} \\
\cmidrule(lr){2-4}
\cmidrule(lr){5-7}
& \textbf{PMR}
& \textbf{AM}
& \textbf{BC}
& \textbf{C10-C}
& \textbf{C100-C}
& \textbf{IN-C} \\
\midrule
ZS
& -
& -
& -
& 61.14
& 35.80
& 24.52 \\

Bias only
& -
& -
& \checkmark
& 67.06
& 40.28
& 28.07 \\

PMR only
& \checkmark
& -
& -
& 58.23
& 35.99
& 33.27 \\

PMR + AM
& \checkmark
& \checkmark
& -
& 77.38
& 51.07
& 34.35 \\

PMR + BC
& \checkmark
& -
& \checkmark
& 59.87
& 38.48
& 35.11 \\

Full
& \checkmark
& \checkmark
& \checkmark
& \textbf{77.91}
& \textbf{53.29}
& \textbf{36.08} \\
\bottomrule
\end{tabular*}
\vspace{-0.3cm}
\end{table}

\begin{figure*}[t]
    \centering
    \includegraphics[width=\textwidth]{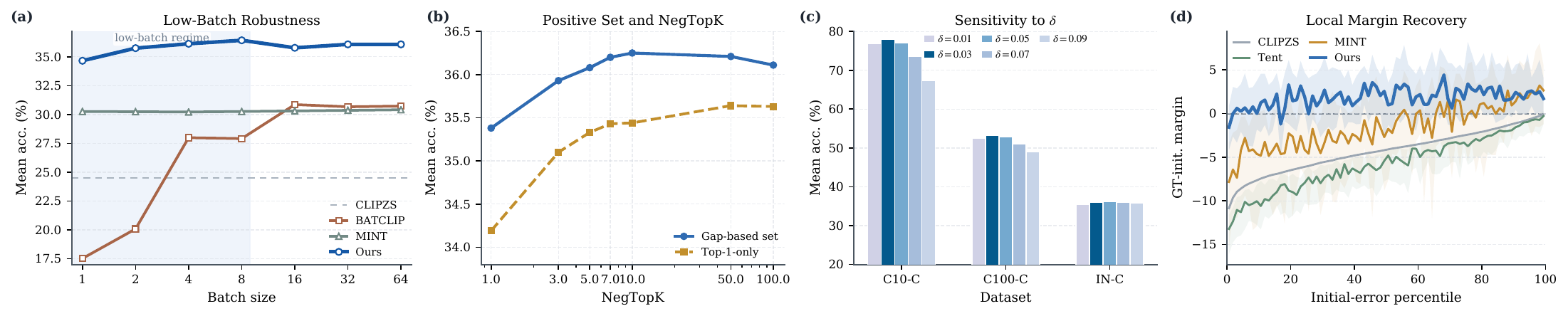}
    \vspace{-0.5cm}
    \caption{Robustness and PMR ablations. 
(a) Our method remains robust across batch sizes, especially in the low-batch regime. 
(b) The gap-based positive set consistently outperforms the top-1-only variant, while performance is relatively insensitive to NegTopK. 
(c) Performance is stable across a moderate range of the protected-gap threshold $\delta$, with the default choice lying inside the stable region.
(d) Our method more effectively restores the local margin on initially misclassified samples.}
\Description{Ablation charts analyzing batch-size robustness, negative top-K sensitivity, protected-gap threshold delta, and margin recovery.}
\vspace{-0.1cm}
    \label{fig:ablation_others}
\end{figure*}

\noindent\textbf{Low-Batch Robustness.}
Figure~\ref{fig:ablation_others}(a) compares our method with MINT and BATCLIP under varying batch sizes on ImageNet-C. Our method consistently performs best across all tested batch sizes, improving from 34.66 at batch size 1 to 36.08 at batch size 64. The advantage is particularly pronounced in the low-batch regime: at batch size 1, our method exceeds MINT and BATCLIP by 4.40 and 17.14 points, respectively. While BATCLIP is highly sensitive to small batches and MINT remains relatively stable but weaker overall, our method reaches a near-saturated performance level with only modest batch sizes, indicating that it does not rely on large batches to produce effective online updates.

\noindent\textbf{Effect of Positive Set and Hard Negatives.}
Figure~\ref{fig:ablation_others}(b) studies two implementation choices in PMR: the positive-set definition and the hard-negative set size. The gap-based positive set consistently outperforms the top-1-only variant for every tested value of NegTopK, indicating that preserving a plausible near-top candidate set is more effective than restoring only the current top-1 prediction. By contrast, performance is relatively stable across different NegTopK values. On ImageNet-C, the gap-based variant stays within a narrow range of 35.38--36.25 and achieves its best result at NegTopK$=10$. Overall, these results suggest that the main gain comes from the protected near-top positive set, while a moderate hard-negative set size is already sufficient in practice.

\noindent\textbf{Hyperparameter Sensitivity.}
Among the hyper-parameters in our method, $\delta$ is the most directly tied to the definition of the protected near-top set in PMR. We therefore analyze its effect in the main text, while reporting the remaining sensitivity results in the appendix. Figure~\ref{fig:ablation_others}(c) examines the sensitivity to the protected-gap threshold $\delta$. Across all three datasets, performance is stable within a moderate range of $\delta$, showing that PMR is not overly sensitive to the precise boundary used to define the protected near-top set. The default setting $\delta=0.03$ lies inside this stable region and achieves the best or near-best performance on all datasets. In contrast, excessively large values of $\delta$ lead to clearer degradation, especially on CIFAR10-C and CIFAR100-C, because the protected set becomes too loose and begins to include classes that are less relevant to local recovery.

\noindent\textbf{Local Margin Recovery Analysis.}
Figure~\ref{fig:ablation_others}(d) further verifies that our method restores the intended local decision geometry. 
For an initially misclassified sample $x_i$, we define the local margin as $m_i = s_{i,y_i} - s_{i,\hat{y}_i}$, where $y_i$ is the ground-truth label and $\hat{y}_i$ denotes the initial CLIPZS prediction. 
A larger value of $m_i$ indicates that the correct class is better recovered relative to the original erroneous winner. 
As shown in the figure, our method consistently shifts this margin upward compared with Tent and MINT across the full percentile range of initially misclassified samples. 
This suggests that PMR improves local recoverability in a targeted manner, rather than merely sharpening the currently predicted class.

% \noindent\textbf{Robustness to diverse templates.}

\noindent\textbf{Efficiency.}
Table~\ref{tab:efficiency} compares the adaptation cost of different methods under a unified single-sample online setting ($\mathrm{bs}=1$). Training-free methods are the lightest, but they are also substantially weaker in the main accuracy results. Among adaptation methods that update visual LayerNorm parameters only, our method is more efficient than MINT while operating in the same trainable-parameter regime. By contrast, methods that adapt both visual and text-side parameters, such as BATCLIP and BITTA, incur substantially larger latency and memory overhead.

% \begin{table}[t]
% \centering
% \caption{Single-sample test-time efficiency on ImageNet-C (severity 5, $bs=1$).}
% \label{tab:efficiency}
% % \small
% \setlength{\tabcolsep}{4pt}
% \begin{tabular}{lccc}
% \toprule
% Method & \shortstack{Time/img\\(ms)} & \shortstack{Peak mem.\\(MB)} & \shortstack{Trainable\\params} \\
% \midrule
% Source  & 11.4  & 570   & 0 \\
% TDA     & 14.0  & 958   & 0 \\
% ECALP   & 21.8  & 958   & 0 \\
% Tent    & 33.8  & 1052  & 41.0K \\
% ETA     & 17.4  & 1052  & 41.0K \\
% BITTA   & 510.6 & 18.0K & 65.5K \\
% MINT    & 56.2  & 1162  & 39.9K \\
% BATCLIP & 510.6 & 18.0K & 65.5K \\
% Ours    & 45.6  & 1162  & 39.9K \\
% \bottomrule
% \end{tabular}
% \end{table}
\begin{table}[t]
\centering
\caption{{Single-sample test-time efficiency on ImageNet-C (severity 5, $bs=1$).}\vspace{-0.3cm}}
\label{tab:efficiency}
\setlength{\tabcolsep}{8pt} % 适当加大列间距，避免拥挤
\resizebox{0.9\linewidth}{!}{
\begin{tabular}{lccc} % 如果想让数字更整齐，可以考虑把 lccc 改为 lrrr
\toprule
\textbf{Method} & \makecell[c]{\textbf{Time/img}\\\textbf{(ms)}} & \makecell[c]{\textbf{Peak mem.}\\\textbf{(MB)}} & \makecell[c]{\textbf{Trainable}\\\textbf{params}} \\
\midrule
Source  & 11.4  & 570   & 0 \\
TDA     & 14.0  & 958   & 0 \\
ECALP   & 21.8  & 958   & 0 \\
Tent    & 33.8  & 1052  & 41.0K \\
ETA     & 17.4  & 1052  & 41.0K \\
BITTA   & 510.6 & 18.0K & 65.5K \\
MINT    & 56.2  & 1162  & 39.9K \\
BATCLIP & 510.6 & 18.0K & 65.5K \\
Ours    & 45.6  & 1162  & 39.9K \\
\bottomrule
\end{tabular}
}
\vspace{-0.3cm}
\end{table}

% \noindent\textbf{Additional Experiments.}
% To further evaluate the generality of our method, we provide additional experiments under diverse prompt templates and different CLIP backbones in the appendix.

\section{Conclusion}

In this paper, we identified a key limitation of existing online test-time adaptation methods for vision-language models: over-committing to the current top-1 prediction can suppress locally recoverable alternatives and gradually destabilize adaptation over long target streams. To address this issue, we proposed the \textbf{LMR} framework. At the sample level, {PMR} preserves plausible near-top candidates and restores their separation from hard negatives. At the stream level, {AM} and {BC} jointly stabilize online adaptation by regulating class-wise update concentration and compensating persistent output bias. Extensive experiments on image corruptions and ImageNet-Variants benchmarks show that {LMR} consistently outperforms strong baselines while remaining robust in challenging low-batch streaming settings. We hope our findings on local margin restoration and stream-level stabilization can inspire future work on robust adaptation for vision-language models.

\begin{acks}
This work was supported by the Guangdong Basic and Applied Basic Research Foundation under Grant No.~2023A1515110077.
\end{acks}

\bibliographystyle{ACM-Reference-Format}
\bibliography{cite}

\newpage
\appendix

% \noindent\textbf{Appendix}
\section*{Appendix}
In this supplementary material, we provide additional details for the main paper:

\begin{itemize}
\item Derivations for Theoretical Insights in Section~\ref{app:theory}.
    % \item Additional implementation details in Section~\ref{app:imple}.
    \item Details of the baseline methods in Section~\ref{app:baseline}.
    % \item More Hyperparameter Sensitivity in Section~\ref{more_hparam}.
    \item Additional Experiments in Section~\ref{app_exp}.
    
\end{itemize}

\section{Derivations for Gradient Insights}
\label{app:theory}

We derive the effective update structure of LMR, with particular emphasis on the PMR objective and the stream-level stabilizers. Throughout, for a fixed sample $i$, the protected set $\mathcal{P}_i$, candidate weights $\alpha_{ic}$, hard-negative set $\mathcal{N}_i$, and target margin $\gamma_i$ enter the derivation as detached quantities; accordingly, gradients are taken only through the remaining differentiable terms.
Since masked top-$K$ selection introduces only finitely many support-switching boundaries, the loss is piecewise smooth, and the gradients below are valid almost everywhere, i.e., away from those measure-zero switching points.

\subsection{Derivation of the PMR Gradient Structure}
\label{app:protected_gradient}

Recall that the positive and negative scores are defined as
\begin{equation}
s_i^{+}
=
\log \sum_{c\in\mathcal{P}_i}\alpha_{ic}\exp(o_{ic}),
\qquad
s_i^{-}
=
\log \left(
\frac{1}{|\mathcal N_i|}
\sum_{c\in\mathcal N_i}\exp(o_{ic})
\right),
\label{eq:app_pos_neg_scores}
\end{equation}
and that the local protected margin is
\begin{equation}
m_i = s_i^{+} - s_i^{-},
\qquad
\mathcal{L}_i^{\mathrm{PMR}}
=
\operatorname{softplus}(\gamma_i - m_i).
\label{eq:app_lmr_loss}
\end{equation}
Let
\begin{equation}
\nu_i = \sigma(\gamma_i - m_i),
\label{eq:app_nu}
\end{equation}
where $\sigma(\cdot)$ is the logistic sigmoid. Since
\begin{equation}
\frac{\partial \operatorname{softplus}(x)}{\partial x}=\sigma(x),
\end{equation}
we obtain
\begin{equation}
\frac{\partial \mathcal{L}_i^{\mathrm{PMR}}}{\partial m_i}
=
-\sigma(\gamma_i-m_i)
=
-\nu_i.
\label{eq:app_dL_dm}
\end{equation}

We next differentiate the positive and negative scores with respect to the logits. For $c\in\mathcal{P}_i$,
\begin{equation}
\frac{\partial s_i^{+}}{\partial o_{ic}}
=
\frac{\alpha_{ic}\exp(o_{ic})}{\sum_{j\in\mathcal{P}_i}\alpha_{ij}\exp(o_{ij})}
\triangleq
\pi_{ic}^{+},
\label{eq:app_dsplus}
\end{equation}
while for $c\notin\mathcal{P}_i$,
\begin{equation}
\frac{\partial s_i^{+}}{\partial o_{ic}}=0.
\label{eq:app_dsplus_zero}
\end{equation}
Similarly, for $c\in\mathcal{N}_i$,
\begin{equation}
\frac{\partial s_i^{-}}{\partial o_{ic}}
=
\frac{\exp(o_{ic})}{\sum_{j\in\mathcal{N}_i}\exp(o_{ij})}
=
\pi_{ic}^{-},
\label{eq:app_dsminus}
\end{equation}
while for $c\notin\mathcal{N}_i$,
\begin{equation}
\frac{\partial s_i^{-}}{\partial o_{ic}}=0.
\label{eq:app_dsminus_zero}
\end{equation}

Since
\begin{equation}
\frac{\partial m_i}{\partial o_{ic}}
=
\frac{\partial s_i^{+}}{\partial o_{ic}}
-
\frac{\partial s_i^{-}}{\partial o_{ic}},
\label{eq:app_dm_do}
\end{equation}
 and the protected set and the hard-negative set are disjoint, combining Eqs.~\eqref{eq:app_dL_dm}--\eqref{eq:app_dm_do} yields
\begin{equation}
\frac{\partial \mathcal{L}_i^{\mathrm{PMR}}}{\partial o_{ic}}
=
\begin{cases}
-\nu_i \pi_{ic}^{+}, & c\in\mathcal{P}_i,\\[3pt]
\phantom{-}\nu_i \pi_{ic}^{-}, & c\in\mathcal{N}_i,\\[3pt]
0, & c\notin \mathcal{P}_i\cup\mathcal{N}_i.
\end{cases}
\label{eq:app_lmr_logit_grad}
\end{equation}

Equation~\eqref{eq:app_lmr_logit_grad} shows that the PMR term is explicitly support-restricted at the logit level: it assigns nonzero gradients only to classes in $\mathcal{P}_i \cup \mathcal{N}_i$. Protected classes receive negative logit gradients, whereas external hard negatives receive positive logit gradients. Under a direct gradient-descent step on the logits, this promotes the protected classes and suppresses the hard negatives. In particular, if the ground-truth class lies in $\mathcal{P}_i$, it is excluded from the negative branch and can only receive reinforcement through the protected positive term.

\paragraph{Top-1 degeneration.}
When $\mathcal{P}_i=\{\hat y_i\}$, only the current top-1 class receives positive-side support. Any recoverable near-top class outside $\mathcal{P}_i$ loses that protection; if it is selected into $\mathcal{N}_i$, it receives explicit negative pressure, and otherwise it receives no direct gradient. This explains why the top-1-only variant can still work, yet remains weaker than the full protected-set formulation when several classes stay close in score.

\subsection{Derivation of the Entropy Gradient}
\label{app:entropy_grad}

For comparison, consider entropy minimization on the predictive distribution
\begin{equation}
\mathcal{L}_i^{\mathrm{ent}}
=
-\sum_c q_{ic}\log q_{ic},
\qquad
q_{ic}=\frac{\exp(o_{ic})}{\sum_j \exp(o_{ij})}.
\label{eq:app_ent_loss}
\end{equation}
Using the softmax derivative
\begin{equation}
\frac{\partial q_{ij}}{\partial o_{ic}}
=
q_{ij}\bigl(\mathbb{I}[j=c]-q_{ic}\bigr),
\label{eq:app_softmax_grad}
\end{equation}
we obtain
\begin{align}
\frac{\partial \mathcal{L}_i^{\mathrm{ent}}}{\partial o_{ic}}
&=
-\sum_j \frac{\partial q_{ij}}{\partial o_{ic}}(\log q_{ij}+1) \\
&=
-\sum_j q_{ij}\bigl(\mathbb{I}[j=c]-q_{ic}\bigr)(\log q_{ij}+1) \\
&=
-q_{ic}(\log q_{ic}+1)
+
q_{ic}\sum_j q_{ij}(\log q_{ij}+1).
\end{align}
Since
\begin{equation}
\mathcal{H}(\mathbf q_i)
=
-\sum_j q_{ij}\log q_{ij},
\qquad
\sum_j q_{ij}=1,
\end{equation}
we have
\begin{equation}
\sum_j q_{ij}(\log q_{ij}+1)
=
-\mathcal{H}(\mathbf q_i)+1.
\end{equation}
Therefore,
\begin{equation}
\frac{\partial \mathcal{L}_i^{\mathrm{ent}}}{\partial o_{ic}}
=
q_{ic}\bigl(-\mathcal{H}(\mathbf q_i)-\log q_{ic}\bigr).
\label{eq:app_entropy_grad_final}
\end{equation}
Unlike Eq.~\eqref{eq:app_lmr_logit_grad}, this gradient is dense over all classes, which explains why entropy minimization globally sharpens the full predictive distribution rather than restoring a local near-top margin.

\subsection{Derivation of Adaptive Margin as Negative Feedback}
\label{app:adaptive_feedback}

Under the same conditioning, the adaptive margin affects the per-sample update only through
\begin{equation}
\nu_i=\sigma(\gamma_i-m_i),
\label{eq:app_nu_again}
\end{equation}
where $m_i$ is treated as fixed with respect to the stream statistic $\rho_i$ in this analysis. With
\begin{equation}
\gamma_i=\tilde\gamma_0-\beta\log(\rho_i+\epsilon),
\label{eq:app_gamma_rho}
\end{equation}
we have
\begin{equation}
\frac{\partial \gamma_i}{\partial \rho_i}
=
-\frac{\beta}{\rho_i+\epsilon}.
\label{eq:app_dgamma_drho}
\end{equation}
Using
\begin{equation}
\frac{\partial \nu_i}{\partial \gamma_i}
=
\nu_i(1-\nu_i),
\label{eq:app_dnu_dgamma}
\end{equation}
the chain rule gives
\begin{equation}
\frac{\partial \nu_i}{\partial \rho_i}
=
\frac{\partial \nu_i}{\partial \gamma_i}
\frac{\partial \gamma_i}{\partial \rho_i}
=
-\frac{\beta}{\rho_i+\epsilon}\nu_i(1-\nu_i).
\label{eq:app_dnu_drho}
\end{equation}
Since $\beta>0$, $\rho_i+\epsilon>0$, and $\nu_i(1-\nu_i)>0$ whenever $0<\nu_i<1$, we obtain
\begin{equation}
\frac{\partial \nu_i}{\partial \rho_i}<0.
\label{eq:app_negative_feedback}
\end{equation}
Hence, the effective restoration pressure decreases monotonically as the running pseudo-class mass grows. This establishes the adaptive margin as a negative-feedback controller against repeated reinforcement of already dominant pseudo-classes.

\paragraph{Confidence-weighted update amplitude.}
If we define the detached confidence-weighted update magnitude as
\begin{equation}
u_i=r_i\nu_i,
\label{eq:app_ui}
\end{equation}
where $r_i$ is the detached confidence gate, then
\begin{equation}
\frac{\partial u_i}{\partial \rho_i}
=
r_i\frac{\partial \nu_i}{\partial \rho_i}
<
0.
\label{eq:app_du_drho}
\end{equation}
Thus, the same monotone negative-feedback property holds for the actual confidence-weighted restoration amplitude.

\subsection{Bias Correction as Relative Logit Reweighting}
\label{app:bias_reweight}

Recall that the corrected logits are
\begin{equation}
\tilde o_{ic}=o_{ic}+b_{t-1,c},
\label{eq:app_tilde_logit}
\end{equation}
and the corresponding softmax probabilities are
\begin{equation}
\tilde q_{ic}
=
\frac{\exp(\tilde o_{ic})}{\sum_j \exp(\tilde o_{ij})}.
\label{eq:app_tilde_softmax}
\end{equation}
Substituting Eq.~\eqref{eq:app_tilde_logit} into Eq.~\eqref{eq:app_tilde_softmax} gives
\begin{equation}
\tilde q_{ic}
=
\frac{\exp(b_{t-1,c})\,\exp(o_{ic})}
{\sum_j \exp(b_{t-1,j})\,\exp(o_{ij})}.
\label{eq:app_bias_reweight}
\end{equation}
Hence, additive correction in logit space is equivalent to multiplicative reweighting in probability space.

This interpretation becomes especially clear from pairwise odds. For any two classes $c$ and $j$,
\begin{equation}
\frac{\tilde q_{ic}}{\tilde q_{ij}}
=
\exp\!\bigl(b_{t-1,c}-b_{t-1,j}\bigr)
\frac{q_{ic}}{q_{ij}},
\label{eq:app_pairwise_odds}
\end{equation}
where $q_{ic}$ denotes the uncorrected softmax probability. Therefore, only relative differences in the bias vector matter.

Now consider the update before projection:
\begin{equation}
\hat{\mathbf b}_t
=
\mathbf b_{t-1}
-
\eta\log(\bar{\mathbf u}_t+\epsilon),
\label{eq:app_hatb}
\end{equation}
where the logarithm is applied element-wise. For any pair of classes $c$ and $j$,
\begin{equation}
\hat b_{t,c}-\hat b_{t,j}
=
(b_{t-1,c}-b_{t-1,j})
-
\eta\log\frac{\bar u_t(c)+\epsilon}{\bar u_t(j)+\epsilon}.
\label{eq:app_pairwise_bias_update}
\end{equation}
If $\bar u_t(c)>\bar u_t(j)$, then the logarithmic ratio is positive, and thus the relative offset for class $c$ decreases. In other words, classes with persistently larger running predictive mass receive stronger negative relative correction.

Finally, the projection
\begin{equation}
P=I-\frac{1}{C}ee^\top
\label{eq:app_projection}
\end{equation}
removes the common mean component. For any vector $\mathbf x$ and any pair $(c,j)$,
\begin{equation}
(P\mathbf x)_c-(P\mathbf x)_j = x_c-x_j.
\label{eq:app_projection_pairwise}
\end{equation}
Thus, the projection discards only the irrelevant global shift, to which the softmax is invariant, while preserving all relative class-wise corrections.

\section{Additional Experimental Details}

% \subsection{Additional Implementation Details}
% \label{app:imple}
% Unless otherwise specified, Adam uses $\beta_1=0.9$, $\beta_2=0.999$, and weight decay $0$. For ImageNet-A, ImageNet-K, ImageNet-V2, and ImageNet-R, we use a learning rate of $1.5\times10^{-3}$; for the other benchmarks, we use $1.0\times10^{-3}$.
% For online bias correction, the residual bias vector is initialized to zero, and the running class-mass estimate is initialized from the first batch, i.e., $\bar{\mathbf{u}}_1=\mathbf{u}_1$. After each update, the bias vector is mean-centered and clipped element-wise to $[-2,2]$. 

\subsection{Baseline implementation details.}
\label{app:baseline}
Unless otherwise specified, we follow the official implementations and default hyperparameter settings of all baseline methods. All baselines are evaluated under the same test-time protocol, using CLIP ViT-B/16 with OpenAI weights. We use the prompt template ``\textit{a photo of a class}.'' for all methods. TDA and ECALP are training-free, while the other baselines update normalization affine parameters online with one adaptation step per batch.

\noindent \textbf{TDA.} We use the training-free TDA setting with batch size 1. We use $\alpha_{\mathrm{pos}}=2.0$, $\beta_{\mathrm{pos}}=5.0$, $\alpha_{\mathrm{neg}}=0.117$, and $\beta_{\mathrm{neg}}=1.0$. We use a normalized entropy range of $(0.2, 0.5)$ for negative-cache selection and a negative mask threshold range of $(0.03, 1.0)$. The positive and negative cache capacities are 3 and 2, respectively.

\noindent \textbf{ECALP.} We use the training-free ECALP setting with batch size 1, $K_{\text{text}}=3$, $K_{\text{image}}=8$, $\gamma=10.0$, $\alpha=1.0$, and 3 label-propagation iterations for all benchmarks.

\noindent \textbf{TENT.} For CIFAR-10-C and CIFAR-100-C, we use Adam with learning rate $2.5\times10^{-4}$, $\beta_1=0.9$, and weight decay 0. For ImageNet-C, ImageNet-A, ImageNet-K, ImageNet-V2, and ImageNet-R, we use SGD with learning rate $2.5\times10^{-4}$ and weight decay 0.

\noindent \textbf{ETA.}  For CIFAR-10-C and CIFAR-100-C, we use Adam with learning rate $2.5\times10^{-4}$, $\beta_1=0.9$, and weight decay 0. For ImageNet-C, ImageNet-A, ImageNet-K, ImageNet-V2, and ImageNet-R, we use SGD with learning rate $2.5\times10^{-4}$ and weight decay 0. We use the default entropy margin $E_0=0.4$, and set the diversity margin to $d=0.05$.

\noindent \textbf{BATCLIP.} We use one adaptation step per batch. For CIFAR-10-C, we use AdamW with learning rate $10^{-3}$, weight decay 0.01, and batch size 200. For CIFAR-100-C, we use Adam with learning rate $5\times10^{-4}$, weight decay 0.01, and batch size 200. For ImageNet-C, ImageNet-A, ImageNet-K, ImageNet-V2, and ImageNet-R, we use AdamW with learning rate $5\times10^{-4}$, weight decay 0.01, and batch size 64. Following the bimodal adaptation setting, the text encoder is not frozen.

\noindent \textbf{MINT.} We use Adam with learning rate $7\times10^{-3}$, $\beta_1=0.9$, weight decay 0, and batch size 64 for all benchmarks. The prior strength for text-weight mixing is set to 10000. Following the original design, the model is reset after each batch while the running accumulators are preserved within the stream.

\noindent \textbf{BITTA.} We implement BITTA on top of BATCLIP by adding bilateral entropy-based sample selection and a high-entropy unlearning term. For CIFAR-10-C, we use AdamW with learning rate $10^{-3}$, weight decay 0.01, and batch size 200. For CIFAR-100-C, ImageNet-C, ImageNet-A, ImageNet-K, ImageNet-V2, and ImageNet-R, we use AdamW with learning rate $5\times10^{-4}$, weight decay 0.01, with batch size 200 on CIFAR-100-C and 64 on the ImageNet-based benchmarks. We set $\alpha=-3.8\times10^{-4}$, $\beta=0.83$, high-entropy ratio $0.1$, and unlearning weight $1.2$. The text encoder is not frozen.

\section{Additional Experiments.}
\label{app_exp}

\noindent\textbf{Robustness on a Larger Backbone.}
To verify that the proposed method is not restricted to a specific backbone, we further evaluate the results on the larger ViT-L/14 backbone on ImageNet-C. We follow the same evaluation protocol as in the main paper and report classification accuracy under 15 corruption types at severity level 5. Table~\ref{tab:imagenet_c_vitl14} shows that the proposed method consistently outperforms all competing methods across all corruption types and achieves the best average accuracy of 50.44\%. This result suggests that the proposed adaptation mechanism remains effective when transferred to a stronger CLIP backbone.

\begin{table*}[htbp]
\centering
\caption{Classification accuracy (\%) on ImageNet-C using the \textbf{ViT-L/14} backbone. We evaluate the robustness of various methods under 15 corruption types at severity level 5. \textbf{Bold} denotes the best results.}
\label{tab:imagenet_c_vitl14}
\setlength{\tabcolsep}{3.8pt} % 针对 16 列数据优化列间距
\resizebox{\textwidth}{!}{
\begin{tabular}{l|ccccccccccccccc|c}
\toprule
\textbf{Method} & \textbf{Gauss.} & \textbf{Shot} & \textbf{Impu.} & \textbf{Defo.} & \textbf{Glas.} & \textbf{Moti.} & \textbf{Zoom} & \textbf{Snow} & \textbf{Fros.} & \textbf{Fog} & \textbf{Brit.} & \textbf{Cont.} & \textbf{Elas.} & \textbf{Pixl.} & \textbf{Jpeg} & \textbf{AVG} \\
\midrule
CLIPZS  & 27.36	& 29.44	& 28.68	& 34.62	& 25.24	& 40.94	& 36.76	& 49.86	& 44.10	& 49.70	& 65.38	& 35.10	& 30.34	& 53.50	& 42.24	& 39.55 \\
TDA     & 27.44	& 29.60	& 29.52	& 34.78	& 26.28	& 41.86	& 37.58	& 50.80 & 45.58 & 51.46	& 65.96	& 36.18	& 31.24	& 54.02	& 42.46	& 40.32 \\
ECALP   & 29.58	& 30.24	& 31.54	& 36.74	& 29.18	& 44.10	& 38.72	& 51.84	& 47.02	& 53.06	& 66.38	& 38.82	& 32.84	& 55.72	& 44.14	& 41.99 \\
TENT    & 32.56	& 33.92	& 33.16	& 36.64	& 31.32	& 42.74	& 38.82	& 50.80	& 44.94	& 51.28	& 65.96	& 42.68	& 34.82	& 54.32	& 47.46	& 42.76 \\
ETA     & 35.88	& 37.50	& 37.18	& 39.12	& 37.22	& 46.2	& 42.10	& 52.26	& 48.10	& 54.40 & 66.72	& 47.36	& 40.42	& 55.90	& 51.72	& 46.14 \\
BATCLIP & 33.44 & 33.54 & 34.44 & 35.36 & 32.50 & 43.48 & 40.82 & 51.00 & 44.44 & 53.22 & 66.16 & 42.72 & 39.54 & 53.90 & 50.52 & 43.67 \\
MINT    & 32.48	& 34.08	& 36.72	& 37.90	& 35.40	& 45.22	& 43.28	& 53.58	& 45.52	& 55.06	& 65.78	& 44.30	& 43.26	& 55.66	& 52.20	& 45.36 \\
\midrule
\textbf{Ours} & \textbf{39.44}	& \textbf{41.04}	& \textbf{41.64}	& \textbf{43.26}	& \textbf{41.98}	& \textbf{48.96}	& \textbf{47.46}	& \textbf{56.30}	& \textbf{51.60}	& \textbf{59.24}	& \textbf{68.26}	& \textbf{52.76}	& \textbf{50.32}	& \textbf{58.56}	& \textbf{55.80}	& \textbf{50.44} \\
\bottomrule
\end{tabular}
}
\end{table*}

\noindent\textbf{Results with CuPL Prompt Templates}
To examine whether the proposed method remains effective under stronger text-side priors, we replace the simple prompt template used in the main paper (``a photo of a {class}'') with CuPL ensemble prompts and reevaluate all compatible methods on ImageNet-C. All other experimental settings remain unchanged. As shown in Table~\ref{tab:template}, our method still achieves the best overall performance, reaching an average accuracy of 38.07\%. Compared with the strongest competing method under CuPL prompts, ETA, our method improves the average accuracy by 5.78 points. Moreover, relative to its simple-template counterpart, our method obtains a further gain of 1.99 points. These results suggest that stronger prompt templates provide additional benefits, but do not remove the need for test-time adaptation; instead, they are complementary to our adaptation mechanism. BATCLIP and BITTA are omitted because their formulations are not directly compatible with multi-template ensembling.

\begin{table*}[htbp]
\centering
\caption{Classification accuracy (\%) on ImageNet-C with CuPL ensemble prompts. All other experimental settings follow the main paper. The best results are highlighted in \textbf{bold}.}
\label{tab:template}
\setlength{\tabcolsep}{3.8pt}
\resizebox{\textwidth}{!}{
\begin{tabular}{l|ccccccccccccccc|c}
\toprule
\textbf{Method} & \textbf{Gauss.} & \textbf{Shot} & \textbf{Impu.} & \textbf{Defo.} & \textbf{Glas.} & \textbf{Moti.} & \textbf{Zoom} & \textbf{Snow} & \textbf{Fros.} & \textbf{Fog} & \textbf{Brit.} & \textbf{Cont.} & \textbf{Elas.} & \textbf{Pixl.} & \textbf{Jpeg} & \textbf{AVG} \\
\midrule
CLIPZS & 13.04 & 14.66 & 13.70 & 24.92 & 16.86 & 26.54 & 24.76 & 33.04 & 33.16 & 37.42 & 57.24 & 19.36 & 14.74 & 34.46 & 36.08 & 26.67 \\
TDA & 13.96 & 15.44 & 14.32 & 25.22 & 16.88 & 26.40 & 25.06 & 34.54 & 33.58 & 37.96 & 57.56 & 19.10 & 15.42 & 34.76 & 36.20 & 27.09 \\
ECALP & 15.40 & 17.00 & 15.50 & 26.60 & 17.94 & 27.24 & 27.10 & 36.26 & 34.92 & 39.90 & 58.04 & 21.02 & 17.46 & 37.12 & 36.66 & 28.54 \\
TENT & 7.20 & 8.90 & 11.10 & 27.20 & 21.10 & 28.92 & 26.10 & 34.40 & 33.58 & 39.26 & 57.60 & 24.98 & 15.86 & 38.08 & 39.12 & 27.56 \\
ETA & 20.64 & 21.56 & 21.64 & 27.66 & 24.36 & 31.50 & 28.92 & 37.42 & 35.44 & 42.40 & 58.44 & 29.06 & 24.16 & 40.20 & 40.92 & 32.29 \\
MINT & 21.52 & 22.84 & 22.60 & 28.66 & 24.06 & 32.74 & 29.72 & 35.86 & 33.22 & 42.24 & 57.76 & 27.60 & 22.62 & 40.16 & 42.10 & 32.25 \\
\midrule
\textbf{Ours} & \textbf{24.36} & \textbf{26.72} & \textbf{26.16} & \textbf{31.88} & \textbf{31.72} & \textbf{37.24} & \textbf{36.00} & \textbf{43.80} & \textbf{40.48} & \textbf{47.64} & \textbf{61.12} & \textbf{36.00} & \textbf{35.60} & \textbf{46.20} & \textbf{46.08} & \textbf{38.07} \\
\bottomrule
\end{tabular}
}
\end{table*}

\end{document}